\def\paperID{7082}
\def\confName{CVPR}
\def\confYear{2025}

\def\paperTitle{Physical Plausibility-aware Trajectory Prediction via Locomotion Embodiment}

\def\authorBlock{
    Hiromu Taketsugu\textsuperscript{1} \qquad
    Takeru Oba\textsuperscript{1} \qquad
    Takahiro Maeda\textsuperscript{1}\\
    Shohei Nobuhara\textsuperscript{2} \qquad
    Norimichi Ukita\textsuperscript{1}\\
    \textsuperscript{1}Toyota Technological Institute, Japan  \qquad
    \textsuperscript{2}Kyoto Institute of Technology, Japan \\
    {\tt\small \{sd23426, sd21502, sd21601\}@toyota-ti.ac.jp,
  nob@kit.ac.jp, ukita@toyota-ti.ac.jp}
}

\newif\ifreview 
\newif\ifarxiv \newcommand{\arxiv}{\arxivtrue}
\newif\ifcamera 
\newif\ifrebuttal 

\arxiv

\pdfoutput=1
\documentclass[10pt,twocolumn,letterpaper]{article}
\ifreview \usepackage[review]{cvpr} \fi
\ifarxiv \usepackage[pagenumbers]{cvpr} \fi
\ifrebuttal \usepackage[rebuttal]{cvpr} \fi
\ifcamera \usepackage{cvpr} \fi

\usepackage{graphicx}	
\usepackage{amsmath}	
\usepackage{amssymb}	
\usepackage{booktabs}
\usepackage{times}
\usepackage{microtype}
\usepackage{epsfig}
\usepackage{caption}
\usepackage{float}
\usepackage{placeins}
\usepackage{color, colortbl}
\usepackage{stfloats}
\usepackage{enumitem}
\usepackage{tabularx}
\usepackage{xstring}
\usepackage{multirow}
\usepackage{xspace}
\usepackage{url}
\usepackage{subcaption}
\usepackage[hang,flushmargin]{footmisc}
\usepackage{bm}
\usepackage{arydshln} 
\usepackage{makecell}

\ifcamera \usepackage[accsupp]{axessibility} \fi

% ...

\newcommand{\ctext}[1]{\raise0.2ex\hbox{\textcircled{\scriptsize{#1}}}}

\ifarxiv  \fi

\newcommand{\red}[1]{\textbf{\textcolor{red}{#1}}}
\newcommand{\blue}[1]{\textbf{\textcolor{blue}{#1}}}

\newcommand{\R}[1]{{%
    \textbf{%
        \ifstrequal{#1}{1}{\textcolor{red}{R#1}}{%
        \ifstrequal{#1}{2}{\textcolor{blue}{R#1}}{%
        \ifstrequal{#1}{3}{\textcolor{magenta}{R#1}}{%
        \ifstrequal{#1}{4}{\textcolor{teal}{R#1}}{%
                           \textcolor{cyan}{R#1}%
        }}}}%
    }%
}}

\usepackage{xr-hyper}

\makeatletter
\newcommand*{\addFileDependency}[1]{
  \typeout{(#1)}
  \@addtofilelist{#1}
  \IfFileExists{#1}{}{\typeout{No file #1.}}
}

\makeatother
\newcommand*{\myexternaldocument}[1]{
    \externaldocument{#1}
    \addFileDependency{#1.tex}
    \addFileDependency{#1.aux}
}

\definecolor{cvprblue}{rgb}{0.21,0.49,0.74}
\usepackage[pagebackref,breaklinks,colorlinks,citecolor=cvprblue]{hyperref}
\usepackage[capitalize]{cleveref}
\crefname{section}{Sec.}{Secs.}
\crefname{table}{Table}{Tables}
\crefname{figure}{Fig.}{Figs.}

\ifarxiv \crefname{appendix}{App.}{Apps.}
\else \crefname{appendix}{Suppl.}{Suppls.} \fi

\unless\ifarxiv \myexternaldocument{_supplementary} \fi

\begin{document}
\title{\paperTitle}
\author{\authorBlock}
\maketitle

\vspace*{-6mm}
\begin{abstract}
Humans can predict future human trajectories even from momentary observations by using human pose-related cues. However, previous Human Trajectory Prediction (HTP) methods leverage the pose cues implicitly, resulting in implausible predictions. To address this, we propose Locomotion Embodiment, a framework that explicitly evaluates the physical plausibility of the predicted trajectory by locomotion generation under the laws of physics.
While the plausibility of locomotion is learned with an indifferentiable physics simulator, it is replaced by our differentiable Locomotion Value function to train an HTP network in a data-driven manner. In particular, our proposed Embodied Locomotion loss is beneficial for efficiently training a stochastic HTP network using multiple heads.
Furthermore, the Locomotion Value filter is proposed to filter out implausible trajectories at inference. Experiments demonstrate that our method enhances even the state-of-the-art HTP methods across diverse datasets and problem settings. Our code is available at: \url{https://github.com/ImIntheMiddle/EmLoco}.
\end{abstract}

\vspace*{-6mm}
\section{Introduction}
\label{sec:intro}

\begin{figure}[tb]
  \centering
  \includegraphics[width=\linewidth]{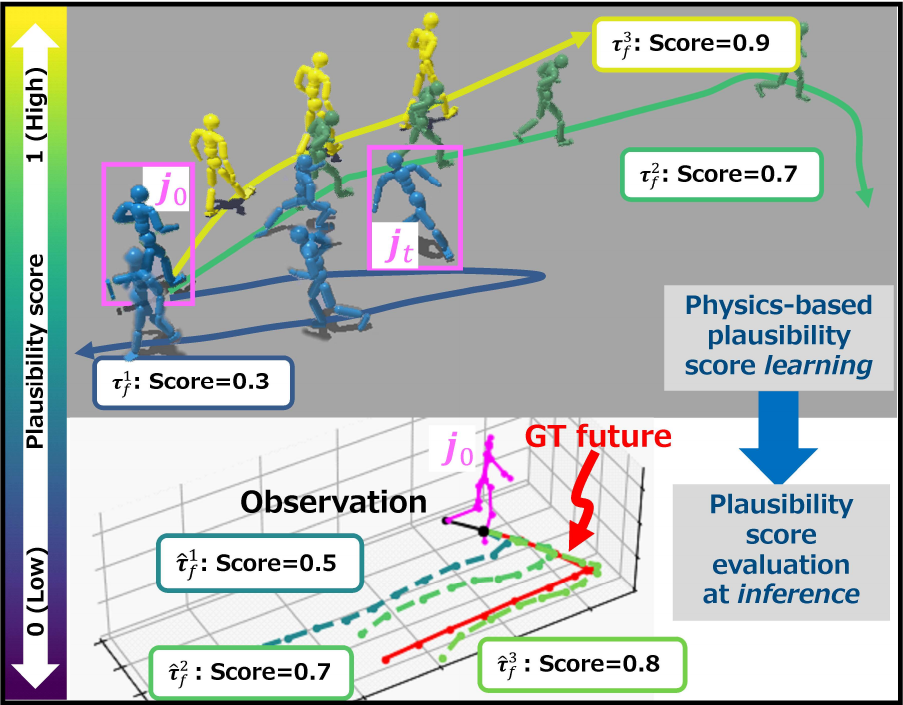}
  \vspace*{-6mm}
  \caption{Overview of our method. Unlike existing methods, which often predict physically implausible trajectories, our framework uses locomotion generation in a physics simulator to incorporate the laws of physics to HTP by training the plausibility {\bf score} as the consistency between the observed 3D pose and future possible trajectories, which are indicated by $\bm{j}_{0}$ and $\bm{\tau}_{f}^{1,2,3}$, respectively, in the figure. Additionally, at inference, our method can evaluate predicted trajectories, $\hat{\bm{\tau}}_{f}^{1,2,3}$, to filter out implausible ones.}
  \label{fig:eye-catch}
\vspace*{-3mm}
\end{figure}

Human Trajectory Prediction (HTP), which predicts future human trajectories,
is actively studied~\cite{SocialLSTM,SocialGAN,Flowchain,ViewBirdiformer,Trajectron++,Eqmotion,MATRIX,CoRLTP_TNT} due to its importance for applications such as autonomous driving~\cite{nuScenes,PacerPlus,HumanDynamicsAD,WhatMatterAD,Waymo,CoRLTP_TNT}, robotics~\cite{JRDB,ViewBirdiformer,HST,ziebart2009planning}, and security systems~\cite{Security,understand_crowd,zhou2015}.
These HTP methods assume the availability of sufficiently long past observations and rely solely on a past trajectory.
Due to this limitation, these methods cannot handle safety-critical situations in a momentary observation setting, where only a few frames of the past trajectory are available, such as when people suddenly emerge from behind obstacles.

Regarding this issue, prior work~\cite{HST,socialtransmotion,quintero2014pedestrian,Pose2019,Next} validates that human pose-related cues, such as
body orientations and
limb motions are beneficial for inferring future trajectories, especially in a momentary observation setting.
These methods learn the consistency between human past poses and future trajectories in a data-driven manner alone.

The data-driven approach can prove its ability with a large-scale dataset that covers most possible combinations of poses and trajectories. However, such datasets are unavailable in the literature because human locomotion is diverse. In particular, it is difficult to collect a large number of accurate pose annotations, which are collected by time-consuming delicate motion capture systems~\cite{AMASS}. 
Instead, we propose to fully leverage the beneficial cues of human poses, by assessing the physical consistency between observed poses and future trajectories for physical plausibility-aware HTP.

For this physical plausibility-aware HTP, we present {\em Locomotion Embodiment}, a framework that incorporates the laws of physics into an HTP network.
This is achieved by physically simulating the locomotion conditioned by observed poses and future trajectories in a physics simulator~\cite{isaac};
refer to Fig.~\ref{fig:eye-catch} to see the effect of our framework.
As a typical example, one of the human poses included in a future possible trajectory $\bm{\tau}_{f}^{1}$ (indicated by $\bm{j}_{t}$ in Fig.~\ref{fig:eye-catch}) is a little unstably tilted, leading to a lower plausibility score of $\bm{\tau}_{f}^{1} = 0.3$.
By training the HTP network in a data-driven manner with following the laws of physics, we benefit from the advantages of both the data-driven and physics-based approaches.

By evaluating and back-propagating the plausibility of locomotion generated in a physics simulator, we can train the physical plausibility-aware HTP network.
However, locomotion generation using physics simulators requires various motion parameters, most of which are unavailable
or noisy (\eg, joint angular velocities estimated from joint positions provided by a 3D pose estimator) in image-based HTP scenarios. 
In addition, rigorous physics simulators are indifferentiable.
To address these issues to back-propagate the physical plausibility, this paper proposes the {\em Locomotion Value (LocoVal) function} implemented with a differentiable network to serve as a surrogate for a physics simulator for the plausibility evaluation (Sec.~\ref{method:value_func}).
Given the plausibility evaluation provided by the LocoVal function, our EmLoco loss penalizes implausible trajectories (Sec.~\ref{method:tp_training}).

This EmLoco loss is used in conjunction with the MSE loss for supervised trajectory prediction learning. While the MSE loss behaves as a data term, our EmLoco loss can be a regularization term that incorporates prior knowledge about physics.
This is crucial for maintaining the diversity of the predicted trajectories.
Although human trajectories are inherently stochastic due to many factors, such as
dynamic environments, training with the MSE loss alone forces all predicted trajectories to be aligned with a single ground truth. That is, minimizing the MSE essentially decreases the diversity of the predicted trajectories.

On the other hand, for stochastically predicting multiple future trajectories, our EmLoco loss allows us to supervise all predicted trajectories to improve their physical plausibility while preserving diversity.
In addition, our LocoVal function can also serve as the {\em LocoVal filter} that evaluates and filters the candidate trajectories at inference (Sec.~\ref{method:filtering}).

Our framework consists of three key components:
\begin{enumerate}
\item {\bf LocoVal function} as a surrogate for an indifferentiable physics simulator in HTP network training.
\item {\bf EmLoco loss} for back-propagating the plausibility of predicted trajectories.
This loss is also useful for deterministic predictors but more beneficial for stochastic predictors because of its applicability to all possible predicted trajectories while preserving their stochastic diversity.
\item {\bf LocoVal filter} for filtering out implausible trajectories at inference. This filter can be applied to any pre-trained HTP network in a plug-and-play manner.
\end{enumerate}
This work is the first attempt to use simulated humanoid control in a physics simulator for HTP.
Since the physics simulator is used only during training, an additional cost (\ie, the cost of the LocoVal function) is negligible at inference.

\section{Related Work}
\label{sec:related}

\subsection{Human Trajectory Prediction (HTP)}
\label{related:tp}

HTP methods can be divided into deterministic~\cite{SocialLSTM,kothari2021human,ViewBirdiformer,Eqmotion,socialtransmotion} and stochastic~\cite{Flowchain,Trajectron++,HST,SocialGAN,MATRIX} methods.
Deterministic HTP methods~\cite{SocialLSTM,kothari2021human,ViewBirdiformer,Eqmotion,SocialForceModel,survey2020} predict the most likely trajectory.
For example, Social-LSTM uses the Long Short Term Memory~\cite{SocialLSTM} to represent the consistency between observed and future trajectories. 
On the other hand, stochastic HTP methods~\cite{Trajectron++,HST,MATRIX,Autobots} produce multiple trajectories or probability distributions. To model the inherent stochasticity in HTP, these methods employ probabilistic frameworks such as Generative Adversarial Networks~\cite{SocialGAN,sophie} and Diffusion models~\cite{MID,LED}, as well as Normalizing Flows~\cite{Flowchain,flomo} to draw multiple trajectories at inference.
Our method is model-agnostic and applicable to both deterministic and stochastic methods.

\subsection{HTP with Pose Information}
\label{related:posetp}

Prior studies demonstrate that human poses benefit HTP ~\cite{socialtransmotion,HST,quintero2014pedestrian,Next,Yagi}.
Yagi~\etal~\cite{Yagi} and Liang~\etal~\cite{Next} incorporate human poses estimated by CNNs into HTP. Transformer-based methods such as Social-Transmotion~\cite{socialtransmotion} and Human Scene Transformer (HST)~\cite{HST} further improve HTP conditioned by human poses.
In particular, HST~\cite{HST} reports the effectiveness of human pose conditioning for momentary observation scenarios.

However, since these methods include poses simply as an additional input to the HTP network, predicted future trajectories can be physically inconsistent with past poses.
Inspired by this observation, our Locomotion Embodiment framework represents the physical plausibility of predicted trajectories as pose-trajectory consistent locomotion.

Note that we aim to improve HTP by fully leveraging human poses, distinct from Human Motion Prediction which prioritizes predicting future poses~\cite{Motionaug,siMLPe,Physmop,LongHMP,Disentangle,TRiPOD}.

\subsection{Locomotion Generation}
\label{related:locogen}

Locomotion generation is a task to generate
walking motions for humanoid robots~\cite{RoboLoco,RSS,Roboloco_SR} and virtual characters~\cite{motionblend,realtime_leg,CGlocomotion_KimL23}.
Locomotion can be controlled on various conditions such as destinations~\cite{WANDR,SAMP,GOAL}, trajectories~\cite{TraceandPace,GMD}, and poses~\cite{PacerPlus,PHC}. Generative models such as
Diffusion models~\cite{MDM,priorMDM,Physdiff,GMD,execute_command} can generate such conditioned locomotion, but they are 
prone to generate physically implausible locomotion (\eg, skating motions).
This is because the laws of physics are not explicitly formulated in generative models.

On the other hand, physics-based locomotion generation~\cite{PacerPlus,TraceandPace,PHC,HumanDynamicsAD,AMP,Physics-controller-cvae} follows the laws of physics in simulators~\cite{isaac,Bullet}.
However, a generated locomotion would fail to follow a given trajectory or fall due to kinematic limitations if a humanoid is controlled to walk along an implausible trajectory.
Therefore, a physics-based locomotion generator can evaluate whether the conditioned trajectory is plausible.
Our Locomotion Embodiment framework utilizes this ability to evaluate the plausibility of future trajectories.

The aforementioned physics-based locomotion generator, \eg, PACER~\cite{PacerPlus,TraceandPace,PHC,AMP}, can be trained with reinforcement learning in a physics simulator.
A straightforward way to use this locomotion generator for training an HTP network is to directly incorporate the locomotion generator into the HTP network.
However, this direct integration has two problems:
(1) As mentioned in Sec.~\ref{sec:intro}, the physics simulator is \textbf{indifferentiable}, preventing training the HTP network by back-propagation.
(2) The locomotion generator using the physics simulator \textbf{requires physical humanoid states}, \eg, joint angular velocities, which are unavailable or noisy in image-based HTP scenarios.
To address these two problems, our framework uses the physics simulator and the locomotion generator only before training the HTP network and replaces them with the LocoVal function in training the HTP network.

\section{Preliminaries}
\label{sec:preliminary}

\paragraph{HTP.}
\label{prelim:htp}
An HTP network, $\mathcal{H}$, takes the past 
$T_{{\text{p}}}$ frames as input, including the past trajectory and 3D poses of a target person, and predicts the trajectory of the target person in the future $T_{\text{f}}$ frames. The past trajectory $\boldsymbol \tau_{\text{p}} 
\in \mathbb{R}^{T_{\text{p}} \times 2}$
and the future target trajectory $\boldsymbol \tau_{\text{f}} \in \mathbb{R}^{T_\text{f} \times 2}$ are on the ground $xy$ plane. The 3D pose involves $J$ joint positions $\boldsymbol{j}_{t} \in \mathbb{R}^{J \times 3}$ at the past time $t$, where $(-T_{\text{p}}+1) \leq t \leq 0$.

The predicted future trajectory $\hat{\boldsymbol \tau}_{\text{f}}$ is expected to be as close as possible to its ground-truth trajectory $\boldsymbol \tau_{\text{f}}$.

\paragraph{Locomotion Generation.}
\label{prelim:locogen}
In our
framework, humanoid locomotion is generated to follow any given future trajectory $\boldsymbol \tau_{\text{s}} \in \mathbb{R}^{T_\text{f}\times2}$ in a physics simulator~\cite{isaac}.
As such a locomotion generator $\mathcal{G}$, our method employs PACER~\cite{TraceandPace}, which is implemented with the policy network~\cite{A2C} trained via reinforcement learning~\cite{PPO}.
$\mathcal{G}$ takes $\boldsymbol \tau_{\text{s}}$ and the initial humanoid state $\bm{h}_{0}$ to control the humanoid joints.
$\bm{h}_{0}$ involves the positions, rotations, velocities, and angular velocities of all joints.
In PACER~\cite{TraceandPace}, $\mathcal{G}$ is trained by maximizing the discounted cumulative reward $\varOmega$ consisting of a trajectory following reward, a motion style reward~\cite{AMP}, and an energy penalty~\cite{fu2022deep}. Refer to PACER~\cite{TraceandPace} for details.

In PACER, the humanoid movement is constrained by the limits of joint angles and torques, preventing it from producing kinematically impossible locomotion.
Due to this limitation, when conditioned on a physically implausible trajectory, the humanoid fails to follow it, resulting in a lower reward.
Therefore, we define the discounted cumulative reward $\varOmega$~\cite{PPO} as the plausibility of $\boldsymbol \tau_{\text{s}}$.


\section{Physical Plausibility-aware HTP}
\label{sec:method}

\begin{figure}[tb]
  \centering
  \includegraphics[width=\linewidth]{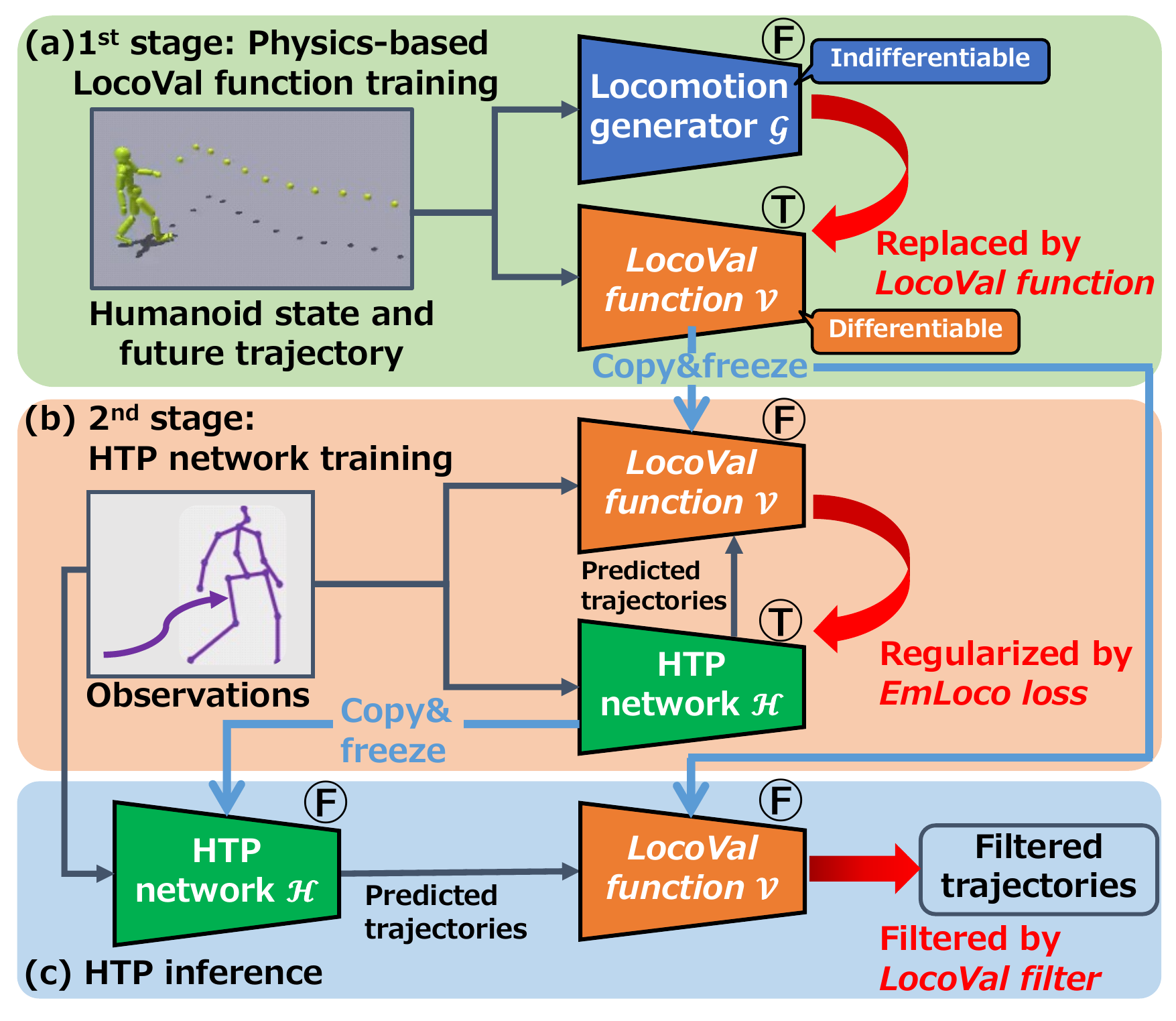}
  \vspace*{-7mm}
  \caption{The overview of the proposed framework.
  \ctext{T} and \ctext{F} mean the weights are trained and fixed, respectively.}
  \label{fig:overview}
\end{figure}

\paragraph{Overview.}
Figure~\ref{fig:overview} shows the overview of our framework consisting of two training stages and an inference phase.
In (a) the {\bf first training stage}, given the physics-based locomotion generator $\mathcal{G}$ trained as mentioned in Sec.~\ref{sec:preliminary}, a current humanoid state is fed into it to obtain the reward as the physical plausibility.
With this physical plausibility as the ground truth, the LocoVal function $\mathcal{V}$ is trained.
In (b) the {\bf second stage}, observed pose-related cues in training data are fed into both $\mathcal{V}$ and $\mathcal{H}$.
$\mathcal{V}$ also takes the future trajectories predicted by $\mathcal{H}$ to compute the EmLoco losses.
The EmLoco loss supports both the MSE and minMSE~\cite{Eqmotion,HST,SocialGAN} losses used for deterministic and stochastic predictions, respectively, to train $\mathcal{H}$.
At (c) {\bf inference}, candidate trajectories predicted by $\mathcal{H}$ are fed into $\mathcal{V}$ to filter out implausible trajectories.


\begin{figure}[tb]
  \centering
  \includegraphics[width=\linewidth]{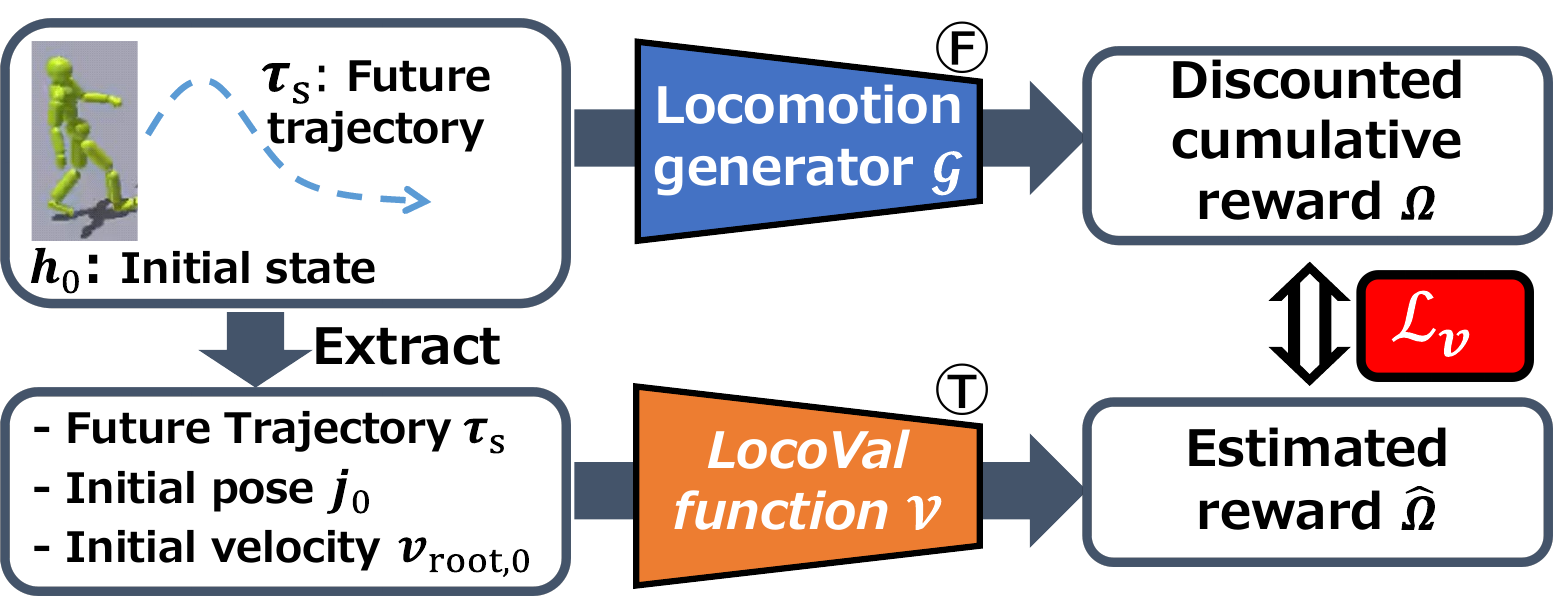}
  \vspace*{-6mm}
  \caption{The overview of training our LocoVal function, $\mathcal{V}$.}
  \label{fig:stage1}
\vspace*{3mm}
  \centering
  \includegraphics[width=\linewidth]{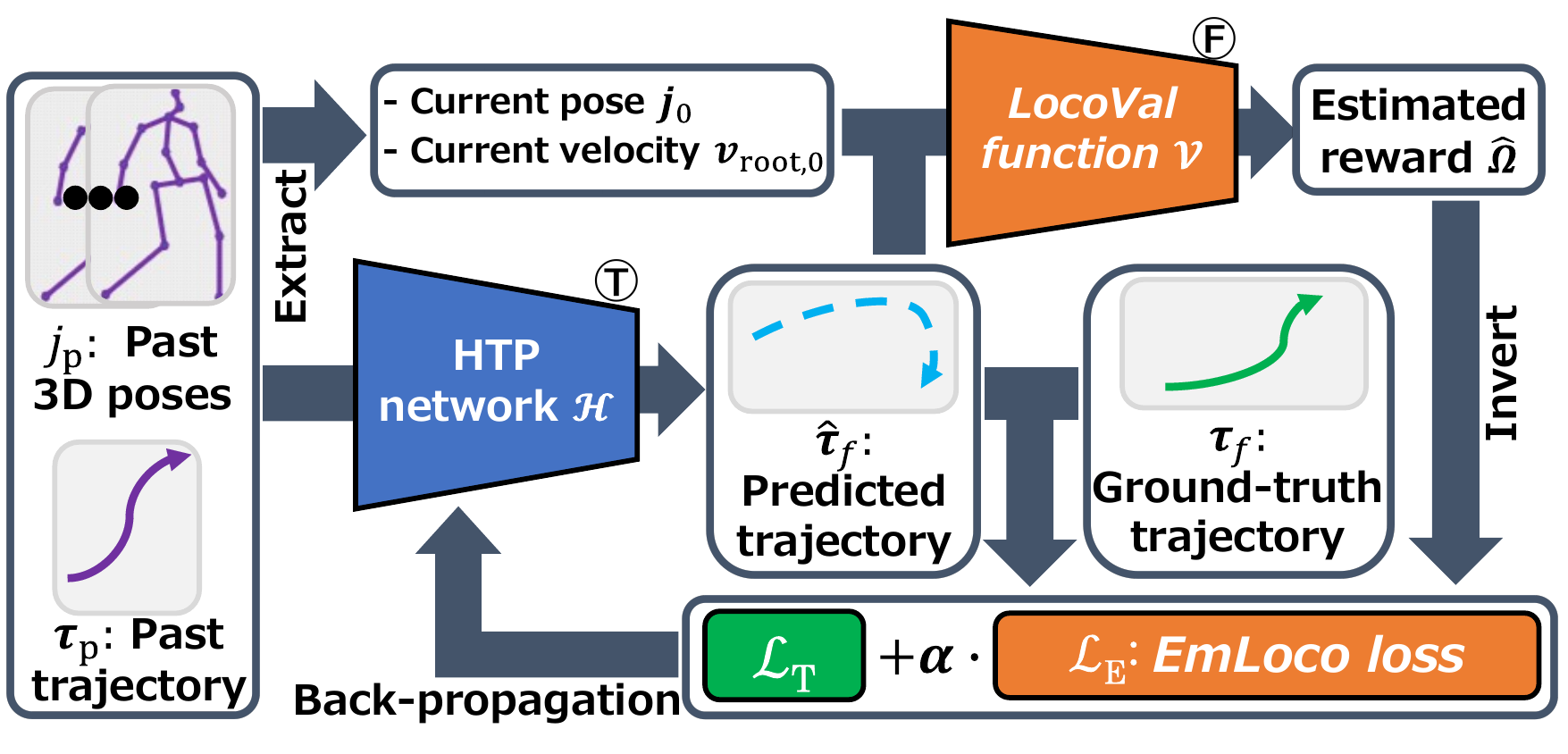}
  \vspace*{-6mm}
  \caption{The overview of training the HTP network, $\mathcal{H}$.}
  \label{fig:stage2}
\end{figure}

\subsection{LocoVal Function}
\label{method:value_func}

The detail of training the LocoVal function $\mathcal{V}$ is illustrated in Fig.~\ref{fig:stage1}. As shown in the top half of Fig.~\ref{fig:stage1}, the discounted cumulative reward $\varOmega$~\cite{PPO} is used as the plausibility of Embodied Locomotion that is generated through $\mathcal{G}$ using a physics simulator.
$\mathcal{V}$ is trained to estimate the reward $\hat{\varOmega}$ close to $\varOmega$ from only available cues in image-based HTP scenarios because $\mathcal{V}$ serves as a surrogate for $\mathcal{G}$ at inference.

As such available cues, our method uses the future trajectory $\boldsymbol \tau_\text{s}$, the initial pose $\bm{j}_{0}$, and the initial velocity of the root joint $\bm{v}_{\text{root},0}\in\mathbb{R}^2$, where $\bm{j}_{0}$ and $\bm{v}_{\text{root},0}$ are in a subset ${\bm{h}'_0}$ extracted from $\bm{h}_{0}$.
In training $\mathcal{V}$, these cues are available from the physics simulator.
At inference, $\boldsymbol \tau_\text{s}$ is replaced by $\boldsymbol \tau_\text{f}$,
$\bm{j}_{0}$ can be obtained by a 3D pose estimator~\cite{blazepose,HST}, and
$\bm{v}_{\text{root},0}$ can be obtained from $\bm{\tau}_{p}$.
$\mathcal{V}$ is implemented as a differentiable MLP, taking these cues as input.

\paragraph{Training Process.}
Unlike PACER~\cite{TraceandPace}, which trains $\mathcal{G}$ to obtain high rewards even for implausible trajectories, \eg, dangerously sharp turns, such 
trajectories should result in low rewards in our framework.
This is because humans do not typically walk along such implausible trajectories in daily life.
For this reason, we train $\mathcal{G}$ with only plausible pairs of $\boldsymbol \tau_\text{s}$ and $\bm{h}_{0}$ to give high rewards only to plausible pose-trajectory pairs.
On the other hand, since $\mathcal{V}$ must discriminate between the plausible and implausible pose-trajectory pairs, it is trained with plausible pairs and randomly paired diverse implausible pose-trajectory pairs.

$\mathcal{V}$ is trained by minimizing the following loss $\mathcal{L}_{\mathcal{V}}$ so that it can estimate the plausibility of the Embodied Locomotion $\varOmega = \mathcal{G}({\boldsymbol \tau_\text{s}}, \bm{h}_0)$ obtained by rolling out $\mathcal{G}$:
\begin{equation}
    \label{eq:mse}
    \mathcal{L}_{\mathcal{V}} = \text{MSE}(\mathcal{V}({\boldsymbol \tau_\text{s}}, {\bm{h}'_0}), \mathcal{G}({\boldsymbol \tau_\text{s}}, \bm{h}_0)),
\end{equation}
where MSE denotes the Mean Squared Error function.

\subsection{HTP Network Training with EmLoco Loss}
\label{method:tp_training}

Figure~\ref{fig:stage2} illustrates the detail of training the HTP network $\mathcal{H}$. Our HTP network takes the past information including $\boldsymbol \tau_{\text{p}}$ and 3D poses $j_{\text{p}}=\{\bm{j}_{-T_\text{p}+1}, \cdots, \bm{j}_{0}\}$ 
as input
to predict the future trajectory $\hat{\boldsymbol \tau}_{\text{f}}$ of a target person. Following traditional HTP training methods~\cite{socialtransmotion,SocialLSTM}, $\hat{\boldsymbol \tau}_{\text{f}}$ is compared with its ground truth $\boldsymbol \tau_{\text{f}}$ to back-propagate the MSE loss $\mathcal{L}_{\text{T}} = \text{MSE}(\hat{\boldsymbol \tau}_{\text{f}}, {\boldsymbol \tau}_{\text{f}})$.

\paragraph{EmLoco loss.}
The EmLoco loss $\mathcal{L}_{\text{E}}$ evaluates predicted trajectories by using $\mathcal{V}$, which has prior knowledge about plausibility.
Note again that a physics simulator is not required in this second stage or the inference phase because $\mathcal{V}$ serves as a surrogate for the plausibility evaluation process.

Given the prediction of $\mathcal{H}$, the predicted trajectory $\hat{\boldsymbol \tau}_{\text{f}}$ and 
the subset of $h_{0}$ (\ie, $h'_{0}$)
observed in the current frame are fed into $\mathcal{V}$ to compute $\mathcal{L}_{\text{E}}$ as follows:
\begin{equation}
    \label{eq:poseloss}
    \mathcal{L}_{\text{E}} = -\mathcal{V}(\hat{\boldsymbol \tau}_{\text{f}}, \bm{h}'_{0}).
\end{equation}
$\mathcal{L}_{\text{E}}$ encourages $\mathcal{H}$ to predict $\hat{\boldsymbol \tau}_{\text{f}}$ that increases the plausibility.

As the total training objective, $\mathcal{L}_{\text{E}}$ is combined with the MSE loss $\mathcal{L}_{\text{T}}$ with a hyper-parameter $\alpha$ as follows:
\begin{equation}
    \label{eq:mse_and_poseloss}
    \mathcal{L}_{\text{total}} = \mathcal{L}_{\text{T}} + \alpha \mathcal{L}_{\text{E}},
\end{equation}

\paragraph{Multi-head HTP network.}
Our method with the EmLoco loss can be extended to a stochastic HTP network using multiple heads.
As also mentioned in Sec.~\ref{sec:intro}, the MSE loss (Eq.~(\ref{eq:mse})) is inappropriate for multiple stochastic predictions because it forces all of them to converge to a unique ground truth.
To mitigate this problem, following prior work~\cite{Eqmotion,HST,SocialGAN}, the multiple heads in our method are trained by replacing $\mathcal{L}_\text{T}$ in Eq.(\ref{eq:mse_and_poseloss}) with the following objective function: 
\begin{equation}
    \label{eq:multi-head}
    \mathcal{L}_{\text{T}} = \min_k(\text{MSE}( \hat{\boldsymbol{\tau}}_{\text{f}}^k,{\boldsymbol{\tau}}_{\text{f}})),
\end{equation}
where $\hat{{\boldsymbol \tau}}_{\text{f}}^k$ denotes the prediction of $k$-th head.
In Eq.~(\ref{eq:multi-head}), the minMSE loss supervises only one prediction closest to its ground truth ${\boldsymbol{\tau}}_{\text{f}}$ among multiple predictions. 

Unlike the MSE loss, this minMSE loss allows us to maintain diverse predictions. However, the minMSE loss is inefficient. 
What is even worse is that several heads may be under-fitted if other heads are trained in early training iterations so that these heads are always selected as those minimizing the MSE in Eq.~(\ref{eq:multi-head}) in later iterations.

On the other hand, our EmLoco loss does not require the unique ground truth, \ie, $\boldsymbol{\tau}_{\text{f}}$ in Eq.~(\ref{eq:multi-head}). As a regularization term, the EmLoco loss trains all the heads to predict plausible trajectories.
This simultaneous optimization allows us to avoid the under-fitting problem mentioned above.
In our method, this simultaneous optimization is implemented by averaging the EmLoco losses using $\mathcal{V}$ for all predicted trajectories, and this average is used as $\mathcal{L}_{\text{E}}$ in Eq.(\ref{eq:mse_and_poseloss}).

\subsection{LocoVal Filter at Inference}
\label{method:filtering}

The trained LocoVal function can also be applied to trajectory filtering, called the LocoVal filter, at inference using multi-head HTP networks.
Similar to the usage of $\mathcal{V}$ for training multiple heads simultaneously at training, it computes the plausibilities of all predicted trajectories at inference.
Each of the predicted trajectories with a plausibility lower than a threshold $\lambda$ is filtered out as follows:
\begin{equation}
        \label{eq:filter}
        \begin{cases} 
        \{ \hat{\boldsymbol{\tau}}_{\text{f}}^k \mid \mathcal{V}(\hat{\boldsymbol{\tau}}_{\text{f}}^k, \bm{h}'_{0}) \geq \lambda \} & \text{if } \exists \, k, \mathcal{V}(\hat{\boldsymbol{\tau}}_{\text{f}}^k, \bm{h}'_{0}) \geq \lambda, \\
         \underset{\hat{\boldsymbol{\tau}}_{\text{f}}^{k}} {\operatorname{argmax}} \, \mathcal{V}(\hat{\boldsymbol{\tau}}_{\text{f}}^k, \bm{h}'_{0}) & \text{otherwise.}
        \end{cases}
\end{equation}

Note that the LocoVal filter can be used with any stochastic HTP networks in a plug-and-play manner.

\section{Experiments}
\label{sec:exp}

In Sec.~\ref{exp:tp_learning}, we benchmark our HTP network trained with the EmLoco loss under the standard HTP setting. Section~\ref{exp:multi} presents results for multi-head HTP networks, and Sec.~\ref{exp:tp_limit_obs} examines the performance with momentary observations. Then, we report ablation results in Sec.~\ref{exp:ablation}. In Sec.~\ref{exp:filter_jta} and Sec.~\ref{exp:eth}, we evaluate the effectiveness of the LocoVal filter in various scenarios.

\subsection{Experimental Settings}
\label{exp:setting}

\paragraph{Datasets.}
As trajectory datasets, the Joint Track Auto (JTA) dataset~\cite{JTA}, the JackRabbot Dataset and Benchmark (JRDB)~\cite{JRDB}, and the ETH/UCY dataset~\cite{ETH,UCY} are used. For the JRDB dataset, a subset provided by the Social-Trans~\cite{socialtransmotion} was used. The ETH/UCY dataset is used only in Sec.~\ref{exp:eth} to evaluate the performance of the LocoVal filter on pre-trained models.

While JTA also provides 3D human poses synchronized with their trajectories, no 3D pose is annotated in JRDB.
Instead, in JRDB, our experiments use 3D poses estimated from images and 2D pose annotations by using~\cite{blazepose, HST}. While the estimated 3D poses are provided by HST~\cite{HST}, misestimated poses are filtered out in our experiments; see the Supp. for the details.
All the 3D poses mentioned above are converted to the SMPL-style poses~\cite{Pose2SMPL} for a fair comparison with the baseline.

While the 3D poses described above are used in the second stage to train the HTP network $\mathcal{H}$, more accurate 3D poses are expected to train $\mathcal{G}$ and $\mathcal{V}$ to learn the physical plausibility of 3D poses.
Such accurate 3D poses are provided by the Archive of Motion Capture as Surface Shapes (AMASS)~\cite{AMASS}.
While accurate 3D poses are available and can be used to compute the trajectories in motion capture datasets such as AMASS, motion capture data do not scale compared with trajectory datasets, as described in Sec.~\ref{sec:intro}.

To avoid this problem, 3D poses in AMASS are combined with trajectories in the trajectory datasets, \ie, JTA and JRDB, as follows.
Plausible and implausible pose-trajectory pairs are required to train $\mathcal{G}$ and $\mathcal{V}$, as described in Sec.~\ref{method:value_func}.
The plausible pairs are prepared by spatially aligning the 3D pose and the trajectories, which are randomly sampled from AMASS and the trajectory datasets, respectively.
This spatial alignment consists of the orientation and velocity alignments.
Given the heading vector of each pedestrian in these two datasets, the heading orientations are aligned between the motion capture pose from AMASS and the trajectory data.
For the velocity alignment, on the other hand, the trajectory data is expanded or contracted to make the vector norms equal between the motion capture pose and the trajectory data.
Refer to the Supp. for detailed training conditions on the aforementioned datasets.

\paragraph{HTP networks.}
Since multiple people are observed in each frame of the aforementioned datasets, all baseline HTP networks of our method take not only the past trajectory of a target person but also those of other people as input. This allows us to predict the target trajectory affected by the dynamic environment. While our method is introduced to take only the target person as input, all people in the scene are fed into $\mathcal{H}$ following the previous methods.
In the evaluations with the JTA and JRDB datasets, Social-Transmotion (Social-Trans), is used as our baseline without any architectural changes. We extend this baseline of the trajectory predictor $\mathcal{H}$ by plugging the EmLoco loss into the training loss. Following the implementation of Social-Trans, past trajectories, 2D and 3D human poses, and 2D and 3D bounding boxes are fed into $\mathcal{H}$ in the JTA dataset~\cite{JTA}, while the past trajectories, 2D bounding boxes, and 3D human poses are used in the JRDB dataset. Note that the 3D human poses are estimated for the JRDB dataset~\cite{JRDB} as mentioned earlier. For the stochastic version of our method, we modify Social-Trans~\cite{socialtransmotion} by parallelizing the prediction heads with different initial weights.
For comparison, all HTP networks predict the future $T_{\text{f}}=12$ frames at $2.5$ fps on the JTA and JRDB datasets. The number of past frames differs depending on the experiment; \eg, $9$ frames in Sec.~\ref{exp:tp_learning} and Sec.~\ref{exp:multi}, and $2$ frames as momentary observations~\cite{momentary} in Sec.~\ref{exp:tp_limit_obs}.

In experiments on the ETH/UCY dataset~\cite{ETH,UCY}, where a bird's-eye view hinders the use of 3D poses, only past trajectories are fed into $\mathcal{H}$. Here, we employ EqMotion~\cite{Eqmotion}, a state-of-the-art stochastic HTP method that only uses past trajectories, as a pre-trained HTP network. Each past trajectory consists of $T_{\text{p}}=8$ frames, while each future trajectory remains at $T_{\text{f}}=12$ frames at $2.5$ fps.

\paragraph{Evaluation.}
Two major metrics, the Average Displacement Error (ADE) and the Final Displacement Error (FDE), both measured in meters, evaluate the HTP results. Consistent with the other HTP literature~\cite{Eqmotion,Trajectron++,Flowchain,HST}, in evaluating the stochastic HTP networks, we report minADE and minFDE among multiple
trajectories. Also, following MATRIX~\cite{MATRIX}, we report $\chi^2$ distances for the four physics primitives of each predicted trajectory, namely velocities, accelerations, angular velocities, and angular accelerations, representing the plausibility of the predicted trajectories.

\begin{table}[tb]
    \caption{HTP results under the standard setting. \red{Red} and \blue{blue} indicate the best and second-best. `Social-Trans' and `Ours' utilize pose input. For methods other than `Social-Trans' and `Ours', we report the results from the Social-Trans literature~\cite{socialtransmotion}.}
    \label{tab:result_tp}
    \begin{center}
        \vspace*{-5mm}
        \begin{tabular}{l|cc|cc}
            \multirow{2}{*}{\textbf{Method}} & \multicolumn{2}{c|}{\textbf{JTA}~\cite{JTA}} & \multicolumn{2}{c}{\textbf{JRDB}~\cite{JRDB}}\\
            & \multicolumn{1}{c}{ADE} & \multicolumn{1}{c|}{FDE} & \multicolumn{1}{c}{ADE} & \multicolumn{1}{c}{FDE} \\ \hline \hline
            Social-GAN-det~\cite{SocialGAN} & 1.66 & 3.76 & 0.50 & 0.99 \\
            Transformer~\cite{giuliari2021transformer} & 1.56 & 3.54 & 0.56 & 1.10 \\
            Vanilla-LSTM~\cite{SocialLSTM} & 1.44 & 3.25 & 0.42 & 0.83 \\
            Occupancy-LSTM~\cite{SocialLSTM} & 1.41 & 3.15 & 0.43 & 0.85 \\
            Directional-LSTM~\cite{kothari2021human} & 1.37 & 3.06 & 0.45 & 0.87 \\
            Dir-social-LSTM~\cite{kothari2021human} & 1.23 & 2.59 & 0.48 & 0.95 \\
            Social-LSTM~\cite{SocialLSTM} & 1.21 & 2.54 & 0.47 & 0.95 \\
            Autobots~\cite{Autobots} & 1.20 & 2.70 & 0.39 & 0.80 \\
            Trajectron++~\cite{Trajectron++} & 1.18 & 2.53 & 0.40 & 0.78 \\
            EqMotion~\cite{Eqmotion} & 1.13 & 2.39 & 0.40 & 0.77 \\ \hline
            Social-Trans~\cite{socialtransmotion} & \blue{1.11} & \blue{2.26} & \blue{0.40} & \blue{0.76} \\ 
            \textbf{Ours} & \red{0.97} & \red{1.91} & \red{0.37} & \red{0.72} \\
        \end{tabular}
        \vspace*{-5mm}
    \end{center}
\end{table}

\subsection{HTP Training with the EmLoco Loss}
\label{exp:emloco}
Here, we evaluate our HTP network trained with the proposed EmLoco loss. Note that the LocoVal filter is not applied in these experiments. $\alpha$ in Eq.(\ref{eq:mse_and_poseloss}) is set to $100$.

\subsubsection{Comparison with Existing HTP Methods}
\label{exp:tp_learning}

Following the standard protocol~\cite{SocialLSTM,socialtransmotion,SocialGAN,Eqmotion,Autobots} using $9$ past frames, 
where past trajectories are sufficiently long,
we compare the accuracy of deterministic predictions by our HTP network with other HTP networks~\cite{SocialGAN, Transformer, SocialLSTM, kothari2021human, Autobots, Trajectron++, Eqmotion, socialtransmotion}.
Table~\ref{tab:result_tp} shows that our method
outperforms all other methods including the state-of-the-art baseline, Social-Trans~\cite{socialtransmotion}, on both the JTA~\cite{JTA} and the JRDB~\cite{JRDB} datasets. This demonstrates that the EmLoco loss effectively provides the prior knowledge of the trajectory-pose consistency during training.
Notably, the performance improvements on the JRDB dataset~\cite{JRDB} with imperfect 3D poses suggest the applicability of our EmLoco loss to real-world applications.


\subsubsection{Multi-head HTP Networks}
\label{exp:multi}

To demonstrate the effectiveness of the EmLoco loss on a multi-head HTP network, $5$ and $20$ prediction heads are implemented on Social-Trans~\cite{socialtransmotion}. As described in Sec.~\ref{method:tp_training}, our proposed method is expected to improve the plausibility of predicted trajectories closer to human locomotion. From this perspective, we evaluate the plausibility of the predicted trajectories by the $\chi^2$ distance employed in MATRIX~\cite{MATRIX}. 

Table~\ref{tab:result_multi} presents the comparisons with the baseline. Our method achieves the best values across almost all metrics on both the JTA~\cite{JTA} and JRDB~\cite{JRDB} datasets. Particularly, the improvements in ADE and FDE reveal that all heads are trained properly because these metrics are evaluated with the prediction of all heads.
Also, improvements in $\chi^2$ distances, the plausibility regarding physics primitives, demonstrate that $\mathcal{H}$ trained with our EmLoco loss predicts trajectories similar to actual human locomotion than the baseline~\cite{socialtransmotion}.

\begin{figure}[tb]
\centering
\includegraphics[width=\linewidth]{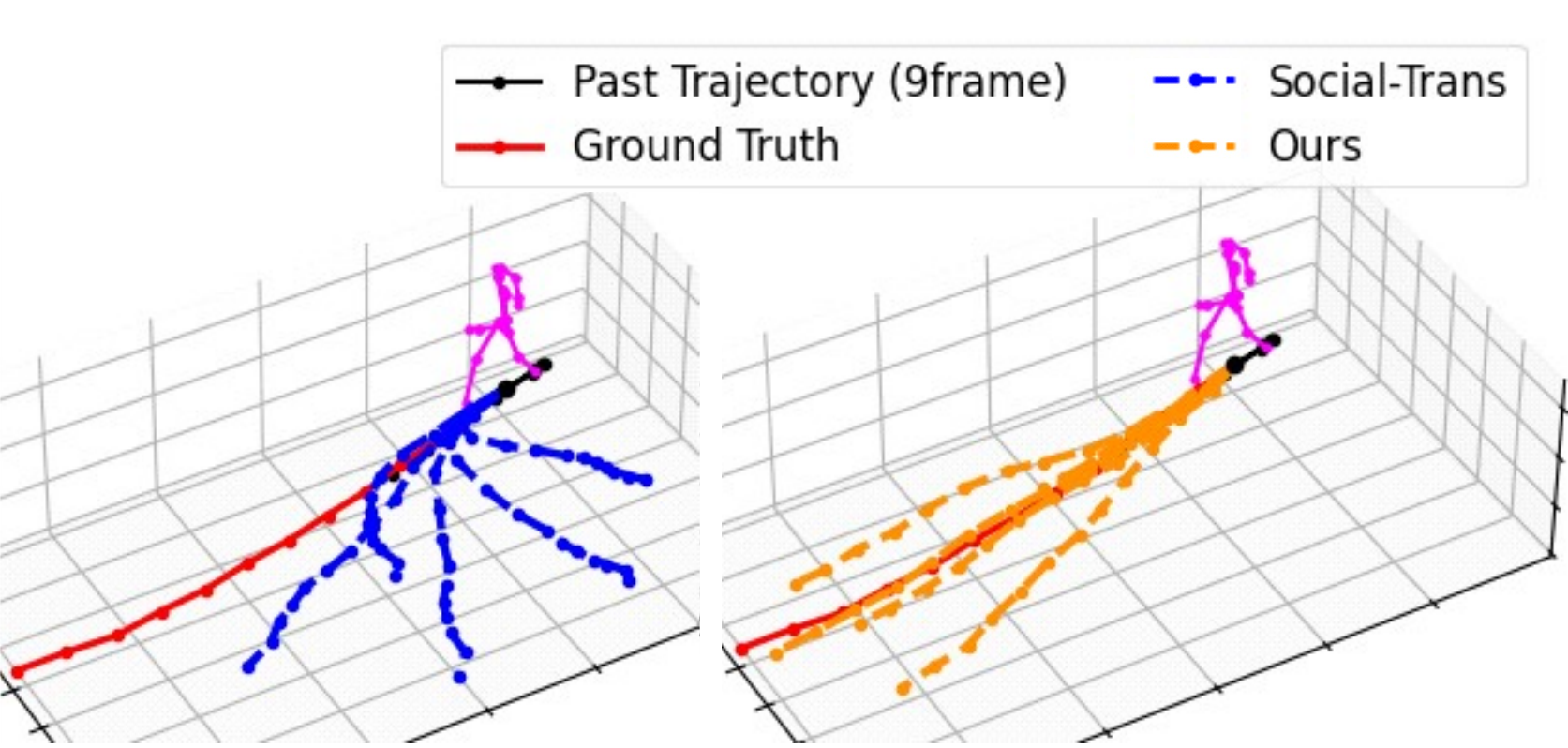}
\caption{Comparison of the stochastic HTP between Social-Trans~\cite{socialtransmotion} and ours with $5$ heads on the JTA dataset~\cite{JTA}. \textbf{\textcolor{magenta}{Human poses}} are shown at a doubled scale for visualization purposes.}
\label{fig:multihead}
\end{figure}

Figure~\ref{fig:multihead} shows qualitative results comparing stochastic HTP with $5$ heads for both Social-Trans~\cite{socialtransmotion} and our method. While the Social-Trans trained solely with minMSE can predict diverse trajectories, many of them are implausible, including sharp turns. In contrast, the predictions of our HTP network maintain diversity across all heads while remaining plausible, resulting in trajectories closer to the ground truth.

\begin{table*}[t]
\caption{Comparison of our HTP network using the EmLoco loss with baseline~\cite{socialtransmotion} in the stochastic HTP under $9$ frames of observations.  \red{Red} indicates the best result for each category. The results are presented as left/right values, where the left denotes evaluations with $5$ heads and the right denotes $20$ heads. `Vel.', `Acc.', `Ang. Vel.', and `Ang. Acc.' represent the $\chi^2$ distances between the distributions of the predicted and ground-truth trajectories for velocity, acceleration, angular velocity, and angular acceleration, respectively.}
\label{tab:result_multi}
\begin{center}
    \vspace*{-5mm}
    \scalebox{0.95}{
    \begin{tabular}{ c | l | c c | c c | c c c c}
                                          & \multirow{2}{*}{\textbf{Method}}       & \multirow{2}{*}{ADE}       & \multirow{2}{*}{FDE}       & \multirow{2}{*}{minADE}    & \multirow{2}{*}{minFDE}    & \multicolumn{4}{c}{$\chi^2$ distances} \\\cline{7-10}
                                    & & & & & & Vel. & Acc. & Ang. Vel. & Ang. Acc. \\ \hline \hline
    \multirow{2}{*}{\small \rotatebox{90}{\textbf{JTA}}}  & Social-Trans & 1.86/2.14 & 3.51/4.26 & 0.88/0.71 & 1.04/\red{0.54} & 0.134/0.169 & 0.009/0.009 & 0.011/\red{0.007} & 0.026/\red{0.008} \\
                                          & \textbf{Ours}         & \red{1.68}/\red{1.80} & \red{3.34}/\red{3.56} & \red{0.79}/\red{0.66} & \red{1.02}/\red{0.54} & \red{0.100}/\red{0.087} & \red{0.002}/\red{0.003} & \red{0.009}/\red{0.007} & \red{0.016}/0.011 \\ \hline
    \multirow{2}{*}{\small \rotatebox{90}{\textbf{JRDB}}} & Social-Trans & 0.61/0.71 & 1.19/1.49 & 0.25/0.18 & \red{0.44}/0.26 & 0.025/0.077 & 0.034/0.013 & \red{0.021}/\red{0.019} & 0.029/0.137 \\
                                          & \textbf{Ours}         & \red{0.56}/\red{0.70} & \red{1.16}/\red{1.48} & \red{0.23}/\red{0.17} & \red{0.44}/\red{0.25} & \red{0.024}/\red{0.067} & \red{0.011}/\red{0.004} & \red{0.021}/0.020 & \red{0.019}/\red{0.036}
    \end{tabular}
    }
    \vspace*{-3mm}
\end{center}
\end{table*}

\subsubsection{HTP with Momentary Observations}
\label{exp:tp_limit_obs}

\begin{table}[tb]
    \caption{HTP results with momentary observations where only $2$ frames of past trajectories are observed. \red{Red} and \blue{blue} indicate the best and second-best for each category. `Traj.' indicates the input is only past trajectories. `Det.' denotes the deterministic prediction.}
    \label{tab:result_momentary}
    \begin{minipage}{0.99\columnwidth}
    \subcaption{HTP with momentary observations on the JTA dataset.}
    \begin{center}
        \vspace*{-5mm}
        \scalebox{0.85}{
        \begin{tabular}{c l|cc|cc}
            \multicolumn{2}{c|}{\textbf{Method}} & ADE & FDE & minADE & minFDE \\ \hline \hline
            \multirow{3}{*}{\rotatebox{90}{\textbf{Det.}}} & Social-Trans (Traj.) & 1.57 & 3.26 & - & - \\
            & Social-Trans      & \blue{1.36} & \blue{2.80} & - & - \\
            & \textbf{Ours}              & \red{1.33} & \red{2.72} & - & - \\ \hline
            \multirow{3}{*}{\rotatebox{90}{\textbf{5heads}}} & Social-Trans (Traj.) & 2.25 & 4.62 & \blue{0.91} & \blue{1.51} \\
            & Social-Trans & \blue{2.23} & \blue{4.55} & 0.92 & 1.53 \\
            & \textbf{Ours}  & \red{1.94} & \red{3.96} & \red{0.87} & \red{1.30} \\ \hline
            \multirow{3}{*}{\rotatebox{90}{\textbf{20heads}}} & Social-Trans (Traj.) & 2.58 & 5.28 & 0.71 & 0.76 \\
            & Social-Trans & \blue{2.46} & \blue{5.07} & \blue{0.69} & \blue{0.73} \\
            & \textbf{Ours}  & \red{2.12} & \red{4.47} & \red{0.65} & \red{0.68} \\
        \end{tabular}
        }
    \end{center}
    \end{minipage}

    \vspace*{3mm}

    \begin{minipage}{0.99\columnwidth}
    \subcaption{HTP with momentary observations on the JRDB dataset.}
    \begin{center}
        \vspace*{-3mm}
        \scalebox{0.85}{
        \begin{tabular}{c l|cc|cc}
            \multicolumn{2}{c|}{\textbf{Method}} & ADE & FDE & minADE & minFDE \\ \hline \hline
            \multirow{3}{*}{\rotatebox{90}{\textbf{Det.}}} & Social-Trans (Traj.) & \blue{0.50} & \blue{0.94} & - & - \\
            & Social-Trans      & \blue{0.50} & \blue{0.94} & - & - \\
            & \textbf{Ours}              & \red{0.44} & \red{0.85} & - & - \\ \hline
            \multirow{3}{*}{\rotatebox{90}{\textbf{5heads}}} & Social-Trans (Traj.) & \blue{0.57} & \blue{1.19} & \blue{0.25} & \red{0.47} \\
            & Social-Trans & 0.58 & 1.20 & \blue{0.25} & \red{0.47} \\
            & \textbf{Ours}  & \red{0.55} & \red{1.12} & \red{0.24} & \blue{0.48} \\ \hline
            \multirow{3}{*}{\rotatebox{90}{\textbf{20heads}}} & Social-Trans (Traj.) & \red{0.68} & \blue{1.46} & \blue{0.18} & \blue{0.30} \\
            & Social-Trans & \red{0.68} & \blue{1.46} & \red{0.17} & \red{0.28} \\
            & \textbf{Ours}  & \red{0.68} & \red{1.45} & \blue{0.18} & \red{0.28} \\
        \end{tabular}
        }
        \vspace*{-3mm}
    \end{center}
    \end{minipage}
\end{table}

As noted in Sec.~\ref{sec:intro}, existing HTP methods struggle with momentary observations. On the contrary, because the proposed EmLoco loss encourages physical plausibility-aware HTP, $\mathcal{H}$ trained with the EmLoco loss is expected to be robust against this setting. From this point of view, we experiment with $2$ past frames following the prior work~\cite{momentary}.

\begin{figure}[tb]
\centering
\includegraphics[width=\linewidth]{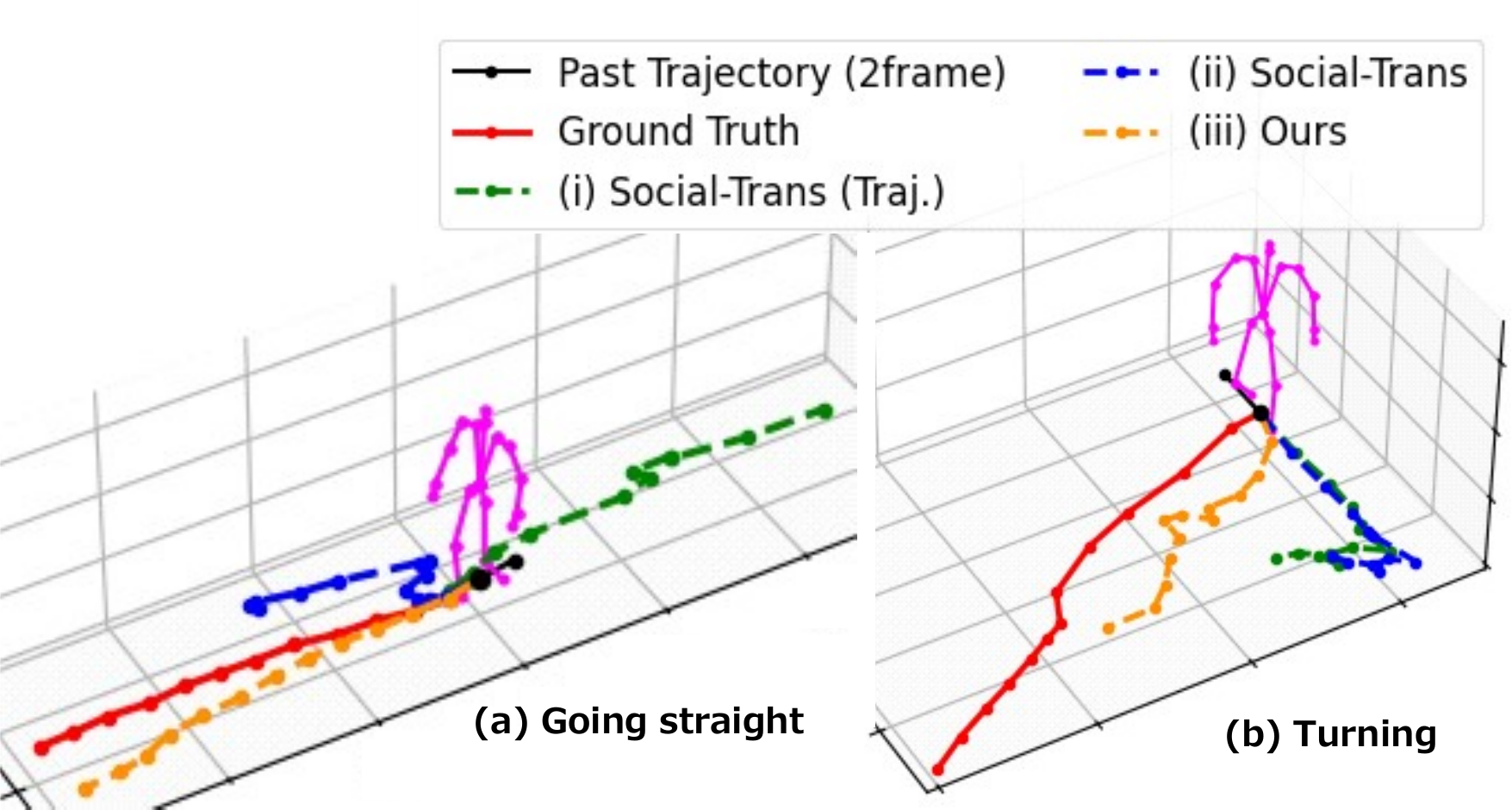}
\caption{Comparison of the prediction results between the baseline Social-Trans~\cite{socialtransmotion} and our method with momentary observations on the JTA dataset~\cite{JTA}. The \textbf{\textcolor{magenta}{observed pose}} are displayed at a doubled scale for visualization purposes.}
\label{fig:limit_obs}
\end{figure}

The results are shown in Table~\ref{tab:result_momentary}. Consistent with Tables~\ref{tab:result_tp} and \ref{tab:result_multi}, 
the EmLoco loss allows our method to outperform the baseline~\cite{socialtransmotion} across almost all categories. 
In addition, Fig.~\ref{fig:limit_obs} presents qualitative results in two cases: when a pedestrian is (a) going straight and (b) turning. In both cases, (i) Social-Trans~\cite{socialtransmotion} with trajectory-only input results in significant prediction errors, and (ii) Social-Trans with pose input still predicts implausible trajectories, especially during sharp turns. In contrast, (iii) our method provides plausible predictions that closely follow the ground-truth trajectory and align with the observed pose.
This demonstrates that the proposed method can effectively leverage human poses and achieve plausibility-aware HTP, thanks to the physics-based LocoVal function training.

\subsubsection{Ablation Studies}
\label{exp:ablation}

\paragraph{Supervision by the EmLoco loss.}
To provide deeper insights into the impact of the EmLoco loss $\mathcal{L}_\text{E}$ on HTP network training, we train $\mathcal{H}$ only with $\mathcal{L}_\text{E}$ \ie, omitting the loss $\mathcal{L}_{\text{T}}$ with the ground-truth trajectories. The results are presented in Table~\ref{tab:ablation}\textcolor{magenta}{(B)}.
Importantly, even the proposed method without $\mathcal{L}_{\text{T}}$ can improve performance from a randomly initialized network (Table~\ref{tab:ablation}\textcolor{magenta}{(A)}).
This demonstrates that the supervision provided by $\mathcal{L}_\text{E}$ contributes to the improvements in HTP performance, while it is designed to be used in conjunction with $\mathcal{L}_\text{T}$. We emphasize that the combination of $\mathcal{L}_\text{T}$ and $\mathcal{L}_\text{E}$ (Table~\ref{tab:ablation}\textcolor{magenta}{(G)}) improves HTP performance
compared to the baseline~\cite{socialtransmotion} (Table~\ref{tab:ablation}\textcolor{magenta}{(C)}).

\paragraph{Inputs into the LocoVal function.}
We ablate the inputs of $\mathcal{V}$ as follows: (1) implausible pose-trajectory pairs used in training $\mathcal{V}$ (Table~\ref{tab:ablation}\textcolor{magenta}{(D)}), (2) the human pose $\bm{j}_0$ at both training and inference (Table~\ref{tab:ablation}\textcolor{magenta}{(E)}), (3)
and the root velocity $\bm{v}_{\text{root},0}$ at both training and inference (Table~\ref{tab:ablation}\textcolor{magenta}{(F)}).
These results indicate that not only plausible but also implausible pose-trajectory pairs are required for training $\mathcal{V}$ to serve as a surrogate for the locomotion generator. Also, both $\bm{j}_0$ and $\bm{v}_{\text{root},0}$ contribute to performance improvements.
Notably, the HTP performance decreases when $\bm{j}_0$ is not used, demonstrating that the pose is important to estimate the plausibility of predicted trajectories.

\begin{table}[tb]  
    \caption{Ablation studies on JTA~\cite{JTA} and JRDB~\cite{JRDB} datasets. \red{Red} and \blue{blue} indicate the best and second-best results for each category. `Initialized' represents the initialized HTP network for reference.}
    \label{tab:ablation}
    \vspace*{-3mm}
    \begin{center}
    \scalebox{0.88}{
        \begin{tabular}{l|cc|cc}
            \multirow{2}{*}{\textbf{Method}} & \multicolumn{2}{c|}{\textbf{JTA}} & \multicolumn{2}{c}{\textbf{JRDB}} \\
            & \multicolumn{1}{c}{ADE} & \multicolumn{1}{c|}{FDE} & \multicolumn{1}{c}{ADE} & \multicolumn{1}{c}{FDE} \\ \hline \hline
            (A) Initialized & 5.07 & 8.92 & 1.31 & 1.87 \\ \hline
            (B) w/o GT loss $\mathcal{L}_{\text{T}}$ & 3.52 & 8.20 & 0.92 & 1.55 \\
            (C) w/o EmLoco loss $\mathcal{L}_{\text{E}}$~\cite{socialtransmotion} & 1.11 & 2.26 & 0.40 & 0.76 \\
            (D) w/o Implausible trajectories & 1.01 & 2.05 & \blue{0.38} & \blue{0.74} \\
            (E) w/o Pose input $\bm{j}_0$ & 1.07 & 2.24 & 0.39 & 0.75 \\
            (F) w/o Root velocity $\bm{v}_{\text{root},0}$ & \blue{0.99} & \blue{2.03} & \blue{0.38} & \blue{0.74} \\ \hline
            (G) \textbf{Ours} & \red{0.97} & \red{1.91} & \red{0.37} & \red{0.72} \\ 
        \end{tabular}
        }
        \vspace*{-3mm}
    \end{center}
\end{table}
\begin{table}[tb]
\caption{Results of the LocoVal filter for stochastic HTP on the JTA~\cite{JTA} dataset with 9 and 2-frame observations. The results are presented as left/right values, where the left denotes evaluations with $5$ heads and the right denotes $20$ heads. \red{Red} indicates the best result. `Rejected' refers to the performance of rejected trajectories.}
\label{tab:result_filter}
\scalebox{0.75}{
\begin{tabular}{cl|cc|cc}
\multicolumn{2}{c|}{\multirow{2}{*}{\textbf{Method}}}                       & \multicolumn{2}{c|}{\textbf{9 frames}}               & \multicolumn{2}{c}{\textbf{2 frames}}                \\ \cline{3-6} 
\multicolumn{2}{c|}{}                                              & \multicolumn{1}{c}{ADE}       & FDE       & \multicolumn{1}{c}{ADE}       & FDE        \\ \hline
\multicolumn{1}{c}{\multirow{3}{*}{\rotatebox{90}{\makecell{Social- \\ Trans \\ (w/o $\mathcal{L}_\text{E}$)}}}} & w/o Filtering & 1.86/2.14 & 3.51/4.26 & 2.23/2.46 & 4.55/5.07  \\ 
            & w/ Filtering  & \red{1.81}/\red{1.97} & \red{3.49}/\red{3.93} & \red{2.06}/\red{2.21} & \red{4.27}/\red{4.56} \\ 
            & Rejected      & 2.37/3.41 & 3.75/6.65 & 3.83/4.32 & 7.16/8.87  \\ \hline
\multicolumn{1}{c}{\multirow{3}{*}{\rotatebox{90}{\makecell{\textbf{Ours} \\ \textbf{(w/ $\mathcal{L}_\text{E}$)}}}}}         & w/o Filtering & 1.68/1.80 & 3.34/3.56 & 1.94/2.12 & 3.96/4.47  \\ 
            & w/ Filtering  & \red{1.65}/\red{1.76} & \red{3.31}/\red{3.52} & \red{1.90}/\red{2.08} & \red{3.88}/\red{4.41}  \\ 
            & Rejected      & 2.32/2.40 & 3.86/4.19 & 3.22/3.25 & 6.18/6.41  \\
\end{tabular}
}
\end{table}


\subsection{LocoVal Filter at Inference}
\label{exp:filter}

\subsubsection{LocoVal Filter on the JTA Dataset}
\label{exp:filter_jta}
Since our LocoVal filter can be applied in a plug-and-play manner to any HTP network, we apply it to Social-Trans~\cite{socialtransmotion} and our method (\ie, Social-Trans trained with $\mathcal{L}_\text{E}$).

Table~\ref{tab:result_filter} presents the filtering results on the JTA dataset~\cite{JTA} with $\lambda$ set to $0.7$ in Eq.(\ref{eq:filter}). We observe that the LocoVal filter enhances performance across all metrics in both Social-Trans and our method.
Notably, although $\mathcal{L}_\text{E}$ using $\mathcal{V}$ is used in training $\mathcal{H}$ in our method, the LocoVal filter using $\mathcal{V}$ can further enhance the performance of our method through filtering at inference.
Table~\ref{tab:result_filter} also shows the evaluation scores of the rejected trajectories.
These scores exhibit substantially poorer performance than the filtered results, indicating that the LocoVal filter rejects implausible, inaccurate trajectories.

\subsubsection{Zero-shot Filtering on the ETH/UCY Dataset}
\label{exp:eth}

Following the aforementioned experiments on the JTA dataset~\cite{JTA}, we further evaluate the LocoVal filter on the ETH/UCY dataset~\cite{ETH,UCY} using another baseline, EqMotion~\cite{Eqmotion}. Note that this dataset is not used for training $\mathcal{V}$, and due to the bird’s-eye-view of the ETH/UCY dataset, we experimented with $\mathcal{V}$ that does not use human poses as input.

Table~\ref{tab:ethucy} presents the results of applying the LocoVal filter under this zero-shot setting, with $\lambda$ set to $0.8$. Although the dataset and human poses are not used for training $\mathcal{V}$, we observe improvements in both ADE and FDE with the application of the LocoVal filter across all $5$ subsets. While the importance of human poses can be observed in the comparison between Table~\ref{tab:ablation}\textcolor{magenta}{(E)} and \textcolor{magenta}{(G)}, Table~\ref{tab:ethucy} suggests that it is even possible to identify implausible trajectories based only on the current root velocity, even without the human poses.

\begin{table*}[tb]
\centering
\caption{Results of zero-shot filtering by the LocoVal filter on the predictions of a pre-trained trajectory predictor~\cite{Eqmotion} on the ETH/UCY dataset~\cite{ETH,UCY}. `Mean' represents the average performance across the $5$ subsets.}
\scalebox{0.95}{
\begin{tabular}{l|cc:cc:cc:cc:cc|cc}
& \multicolumn{2}{c:}{\textbf{ETH}} & \multicolumn{2}{c:}{\textbf{HOTEL}} & \multicolumn{2}{c:}{\textbf{UNIV}} & \multicolumn{2}{c:}{\textbf{ZARA1}} & \multicolumn{2}{c|}{\textbf{ZARA2}} & \multicolumn{2}{c}{\textbf{Mean}} \\ \cline{2-13} 
\multirow{-2}{*}{\textbf{Method}} & \multicolumn{1}{c}{ADE} & \multicolumn{1}{c:}{FDE} & \multicolumn{1}{c}{ADE} & \multicolumn{1}{c:}{FDE} & \multicolumn{1}{c}{ADE} & \multicolumn{1}{c:}{FDE} & \multicolumn{1}{c}{ADE} & \multicolumn{1}{c:}{FDE} & \multicolumn{1}{c}{ADE} & \multicolumn{1}{c|}{FDE} & \multicolumn{1}{c}{ADE} & \multicolumn{1}{c}{FDE} \\ \hline\hline

Pretrained EqMotion~\cite{Eqmotion} & 2.18 & 4.63 & 0.64 & 1.31 & 1.30 & 2.81 & 0.82 & 1.84 & 0.65 & 1.47 & 1.12 & 2.41 \\ \hline
\textbf{Ours (w/ LocoVal filter)} & \red{1.41} & \red{2.88} & \red{0.61} & \red{1.26} & \red{0.93} & \red{2.04} & \red{0.80} & \red{1.80} & \red{0.64} & \red{1.45} & \red{0.88} & \red{1.89} \\ \hline
Rejected trajectories & 8.89 & 19.72 & 2.69 & 5.53 & 4.33 & 9.18 & 1.70 & 3.67 & 2.21 & 4.72 & 3.96 & 8.56 \\
\end{tabular}
}
\label{tab:ethucy}
\end{table*}

\section{Discussion}
\label{sec:discussion}
\textbf{Analysis of Estimated Plausibility.}
We examine the plausibility estimated by $\mathcal{V}$ and the actual reward in the locomotion generator. To test this, $200$ episodes of locomotion are generated from randomly sampled trajectories and initial humanoid states. The correlation coefficient between the estimated plausibilities and the actual rewards is $\textbf{0.85}$. This demonstrates that $\mathcal{V}$ can serve as a surrogate for the locomotion generator, contributing to performance improvements.

\paragraph{Cost-benefit Analysis.}
The additional training time compared to the baseline~\cite{socialtransmotion} is $\textbf{10}$ \textbf{hours} in the first stage on a single NVIDIA RTX A6000 GPU. The additional cost in the second stage is nearly negligible because only the computation of $\mathcal{V}$ is added, which is a shallow MLP. Importantly, $\mathcal{V}$ can be used with any HTP network, which means only the HTP network needs training in the second stage. Considering these benefits, despite the potential for further cost reductions, our method is highly useful.

\paragraph{Infeasible Trajectory in Dataset.}
Some trajectories in the synthetic JTA dataset~\cite{JTA} are infeasible rather than implausible for a humanoid and result in low rewards. However, since our target is plausible locomotion, like most pedestrians' locomotion in daily scenes, it is reasonable not to reproduce such infeasible trajectories. During HTP training, these infeasible trajectories are deprioritized since they have low plausibilities and the HTP network prioritizes more plausible trajectories instead. This weighting mechanism can be one reason for the performance improvement of the proposed method, as shown in Tables~\ref{tab:result_tp}, \ref{tab:result_multi}, and\ref{tab:result_momentary}.

\paragraph{Limitations and Future Work.}
The smaller performance improvement on the JRDB dataset~\cite{JRDB} compared to the JTA dataset~\cite{JTA} (especially in Table~\ref{tab:result_momentary}) is considered to be derived from the assumption of accurate 3D human poses, which is one limitation of the proposed method.
In this perspective, future research may extend the EmLoco loss to account for the uncertainty of human poses~\cite{VL4Pose, VATL4Pose, KNOWN}.

\section{Concluding Remarks}
\label{sec:conclusion}

This paper presented Locomotion Embodiment, a framework that evaluates the plausibility of predicted trajectories as human locomotion.
Our experiments demonstrated that the proposed EmLoco loss improves HTP performance even with the state-of-the-art HTP method~\cite{socialtransmotion}, without any inference speed deceleration. Furthermore, we established the use of the LocoVal function for trajectory filtering.

Importantly, our EmLoco loss improved HTP performance with limited observations even in the real-world dataset~\cite{JRDB}, demonstrating its effectiveness in real-world scenarios. We hope our findings inspire future physics-aware HTP methods for safe real-world applications.
\clearpage
{\small
\bibliographystyle{ieeenat_fullname}
\bibliography{arxiv_main}
}

\clearpage


\appendix

\renewcommand{\thesection}{\Alph{section}}
\newcommand{\appendixhead}%
{\textbf{\huge Appendix}
\vspace{0.25in}}

\appendixhead

\begin{abstract}
This supplementary material covers details and additional results that could not be included in the main manuscript due to page limitations.
First, Sec.~\ref{sec:imple} describes the implementation details and experimental setup.
Section~\ref{appx:results_JTA} presents additional results with EmLoco loss, including results on another real-world dataset, additional analyses, and more qualitative evaluation.
Section~\ref{sec:locoval_add} provides further evaluations of the LocoVal filter, including results with different HTP network baselines, detailed analyses, and visualizations.
Lastly, Sec.~\ref{appx:failure} discusses failure cases and possible future directions.
\end{abstract}

\section{Implementation Details}
\label{sec:imple}

\subsection{3D Pose Conversion}
\label{appx:pose2smpl}
We adopt the Skinned Multi-Person Linear model (SMPL)~\cite{SMPL} as the pose format because the humanoid for the locomotion generator~\cite{TraceandPace} is designed for the SMPL format. However, the HTP datasets used in our experiments (\ie, JTA~\cite{JTA} and JRDB~\cite{JRDB}) do not employ the SMPL format. 
%
Since the LocoVal function is trained through this humanoid in Stage 1 (explained in Sec.~\ref{method:value_func} in the main paper), it is necessary to align the pose formats of these HTP datasets with SMPL.

\begin{figure}[bt]
    \centering
  \includegraphics[width=\linewidth]{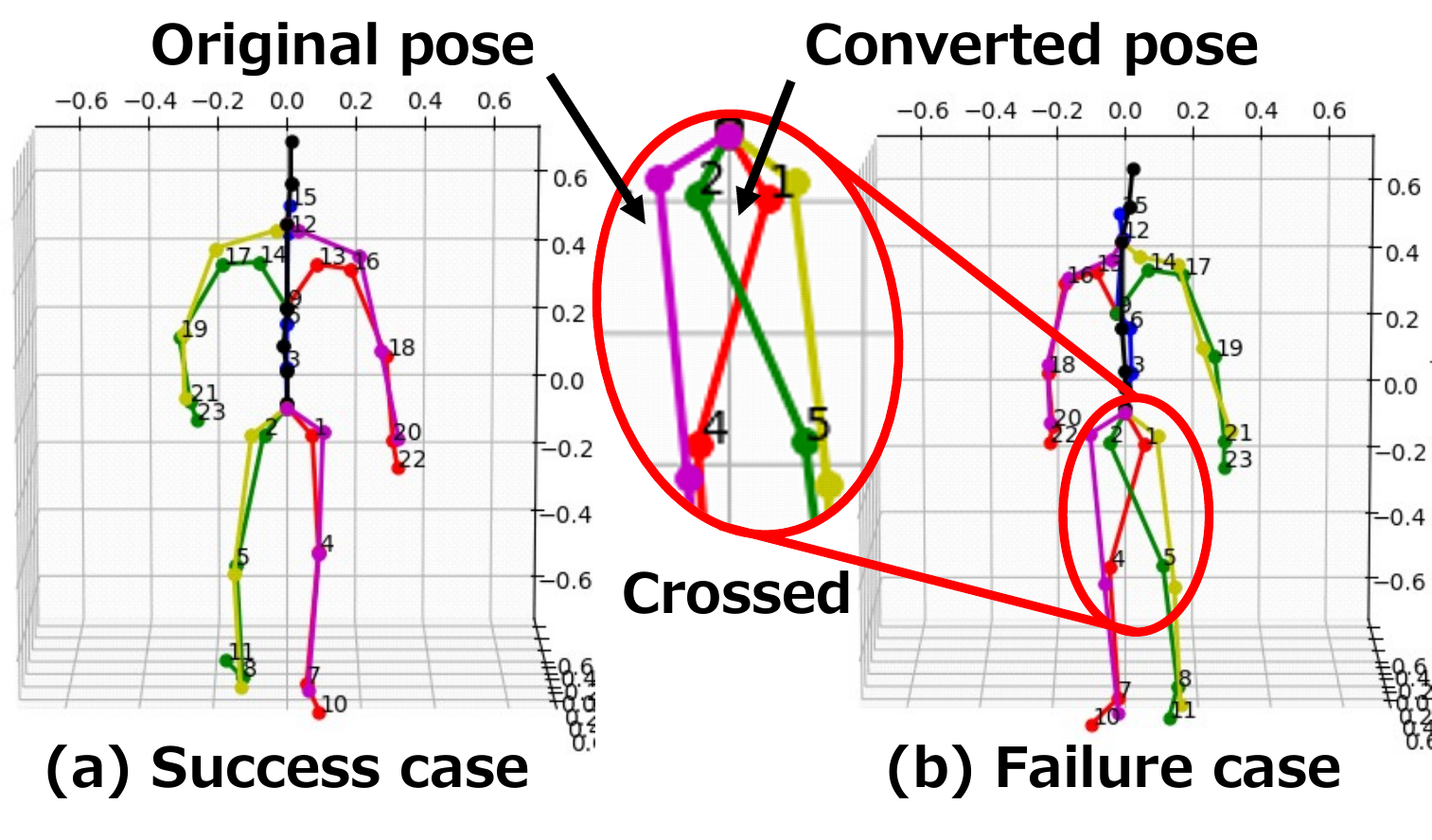}
  \caption{Examples of (a) a success case and (b) a failure case of conversion from the 3D poses in datasets~\cite{JTA,JRDB} to SMPL~\cite{SMPL}. The 3D poses before and after the alignment process are superimposed: the skeletons in {\color[HTML]{DA70D6} \textbf{magenta}}, {\color[HTML]{DAA520} \textbf{yellow}}, and \textbf{black} are the pose before alignment, while the skeletons in \red{red}, {\color[HTML]{008000} \textbf{green}}, and \blue{blue} are the pose after alignment. The numbers next to the poses indicate the joint IDs in the SMPL~\cite{SMPL}.}
  \label{fig:pose2smpl}
\end{figure}

To align the pose formats of the HTP datasets with SMPL, we utilize Pose to SMPL~\cite{Pose2SMPL} for both the JTA and JRDB datasets. Figure~\ref{fig:pose2smpl} illustrates an example of converting a 3D pose from the JTA format into the SMPL format. In Fig.~\ref{fig:pose2smpl}, the poses before and after the alignment process are superimposed: the skeletons in magenta, yellow, and black are the poses before alignment, while the skeletons in red, green, and blue are the poses after alignment. As shown in Fig.~\ref{fig:pose2smpl}(a), this alignment process preserves the original pose, enabling consistent format conversion without significant deformation.

However, we observed that joints are sometimes swapped after applying Pose to SMPL~\cite{Pose2SMPL}. For instance, Fig.~\ref{fig:pose2smpl}(b) shows an incorrect alignment where the left and right hips are swapped. 


To address this issue, these incorrectly aligned joints are automatically swapped by finding these joints based on the inconsistent configuration of the left and right joints between the parent and child nodes. The aligned poses will be provided in our codebase; further details can be found there. The alignment process is also applied to the test data on the JTA and JRDB datasets.

\subsection{Experimental Setup}
\label{appx:setup}

\paragraph{Locomotion Generator.}
The policy network of the locomotion generator is trained by Advantage Actor-Critic (A2C)~\cite{A2C} using Proximal Policy Optimization (PPO)~\cite{PPO}. The learning rate is $2 \times 10^{-5}$ and Adam~\cite{adam} is used as the optimizer. $1,600$ agents are trained in parallel for $5,000$ episodes. Other settings closely follow PACER~\cite{TraceandPace}; \eg, the reward structure of PACER~\cite{TraceandPace} is used as is.

\paragraph{Locomotion Value (LocoVal) Function.}
Our LocoVal function is trained with $160$ agents in parallel for $25,000$ episodes, and optimized by AdamW~\cite{adamw} with the learning rate of $1 \times 10^{-3}$ and cosine annealing~\cite{cosineannealing}. To diversify the pose-trajectory pairs, trajectories generated by PACER~\cite{TraceandPace} are used in addition to trajectories from the trajectory datasets~\cite{JTA, JRDB}. Future trajectories are extracted at $2.5$ fps. The humanoid's initial state $\bm{h}_{0}$ is sampled from AMASS~\cite{AMASS}.


To stabilize the training, we apply coordinate transformations to the inputs of the LocoVal function by constraining the input space. Specifically, we translate the humanoid’s initial position to the origin and align the orientation of the humanoid and the future trajectory at the current timestep by yaw rotation.
During this transformation, the relative angles between the future trajectory $\boldsymbol{\tau}_\text{f}$, the initial pose $\bm{j_0}$, and the initial root velocity $\bm{v_{\text{root}, 0}}$ are preserved.

\paragraph{HTP Network.}
Our HTP network and the baseline, Social-Transmotion (Social-Trans)~\cite{socialtransmotion}, are trained with all the available annotations on the JTA~\cite{JTA} and JRDB~\cite{JRDB} datasets for $30$ and $100$ epochs, respectively. The learning rate is $1 \times 10^{-4}$. After the training, the trained HTP networks with the best performance on the validation set are selected for evaluation on the test set.

Following Social-Trans~\cite{socialtransmotion}, we applied random masking to train the HTP networks. Specifically, the inputs to the HTP networks are randomly masked in modality level, pose keypoint level, frame level, and location level.
This modality-level masking allows the HTP networks to take
the arbitrary combination of available modalities.
Similarly, the frame-level masking allows the HTP networks to work with an arbitrary number of input past frames. In the experiments with momentary observations, we mask all modalities except for the most recent two frames.

In terms of our Embodied Locomotion (EmLoco) loss, the loss function shown in Eq.~(\ref{eq:poseloss}) is implemented as the Mean Squared Error (MSE) with the maximum plausibility score output from a sigmoid function \ie, $1$.
The weight $\alpha$ of the EmLoco loss in Eq.~(\ref{eq:mse_and_poseloss}) is set to $100$ in the results of all experiments shown in the main manuscript and this supplementary material except for Table~\ref{tab:hyperparameter} in Sec.~\ref{sec:emloco_add}.

\subsection{3D Pose Filtering}
\label{appx:filtering}

While JTA dataset~\cite{JTA} contains the ground truth 3D poses of locomotion in a simulated environment, in-the-wild 3D poses in JRDB dataset~\cite{JRDB} include incorrect estimation and non-locomotion poses (\eg, sitting or lying). Since the proposed method does not account for such incorrect 3D poses or poses other than locomotion, these poses are filtered out from the dataset. We applied three types of filtering as follows:

\vspace{2mm}
\begin{itemize}
    \item \textbf{Rule-based Filtering}:
        We assume that the z-coordinate of the head should be higher than those of the knees and pelvis, and the pelvis should be higher than the ankles and lower than the shoulders when the pedestrian is walking.
        Based on these assumptions regarding the z-coordinate relationships of the joints, our filter removes poses that do not meet these assumptions.
        
    \item \textbf{Consistency-based Filtering}:
        For each pedestrian's pose sequence, this filter removes poses with z-scores greater than $2$ based on the $L2$ distances from the moving average of each joint in the pose sequence.

    \item \textbf{Action-based Filtering}:
        Based on the pose-based labels (\eg, walking and standing) provided in JTA-Act~\cite{JRDBAct}, this filter removes poses labeled with non-locomotion-related actions such as sitting or lying.
\end{itemize}
\vspace{2mm}

\begin{figure}[tb]
    \centering
    \includegraphics[width=\linewidth]{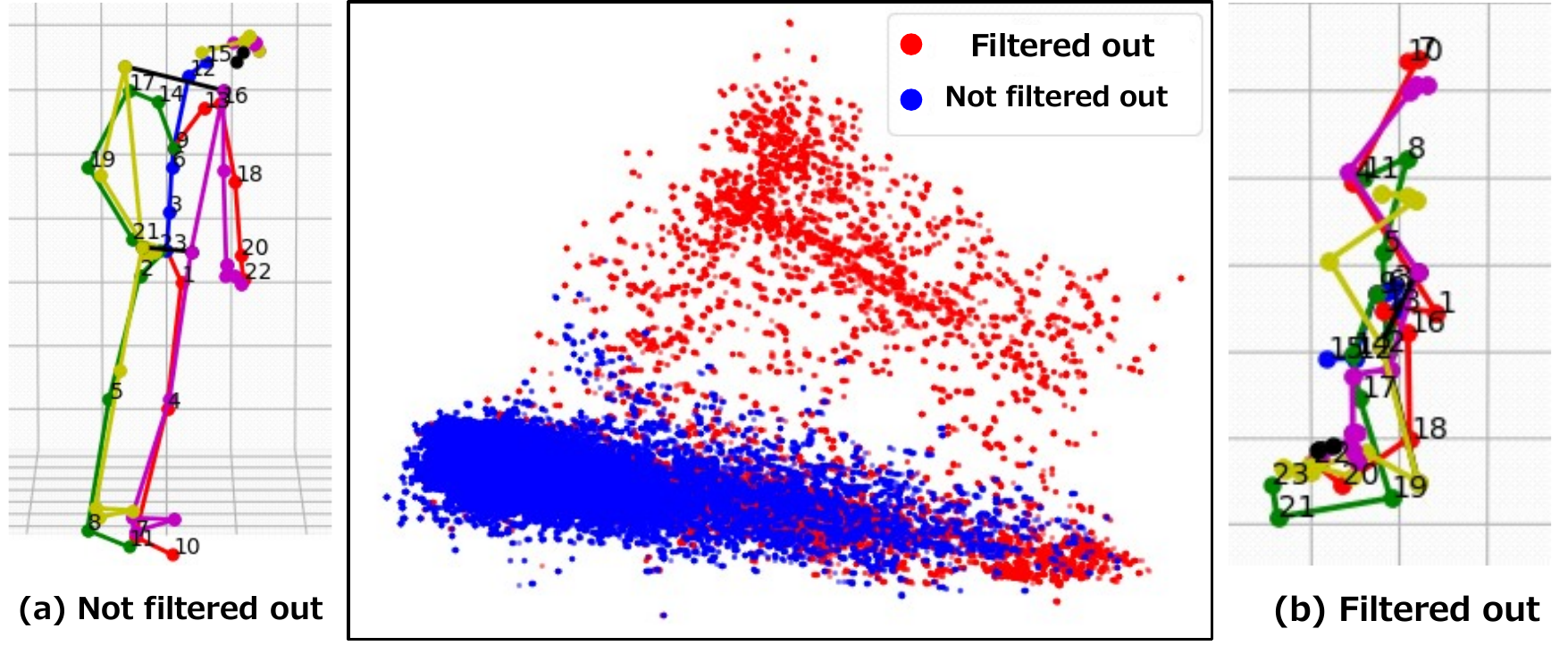}
    \caption{PCA visualization of 3D pose filtering on the JRDB dataset~\cite{JRDB}. \textbf{\textcolor{red}{the red points}} represent the filtered-out data and \textbf{\textcolor{blue}{The blue points}} represent the data that is not filtered out. (a) and (b) are examples of poses that are not filtered out and filtered out, respectively.}
    \label{fig:filter}
\end{figure}

By using these filters, $29.7/17.0/20.7$\% of the 3D poses in the training/validation/test splits are filtered out.
The visualization of Principal Component Analysis (PCA) to the 3D poses treated as vectors is shown in Fig.~\ref{fig:filter}. One can see that \blue{blue points} that are not filtered out are distributed densely. In contrast, \red{red points} that are filtered out are distributed more sparsely over a broader area, indicating poses that deviate significantly from typical locomotion. Examples of pose samples that are not filtered out and filtered out are shown on the left and right of Fig.~\ref{fig:filter}.

Note that this 3D pose filtering affects only the results of our HTP network and Social-Trans, which use the 3D pose as input, while other traditional HTP methods do not require the 3D poses.

\begin{table}[tb]  
    \caption{Ablation studies on different EmLoco loss weights ($\alpha$ in Eq.(\ref{eq:mse_and_poseloss}) of the main paper). Results on the JTA dataset with 9 frames of past observations are shown.}
    \label{tab:hyperparameter}
    \begin{center}
        \begin{tabular}{l|cc|cc}
            \multirow{3}{*}{\textbf{Method}} & \multirow{3}{*}{$\mathcal{L}_\text{T}$} & \multirow{3}{*}{$\mathcal{L}_\text{E}$} & \multicolumn{2}{c}{\textbf{JTA Dataset~\cite{JTA}}} \\
            & & & \multicolumn{2}{c}{\textbf{9 frames}}\\
            & & & \multicolumn{1}{c}{ADE $\downarrow$} & \multicolumn{1}{c}{FDE $\downarrow$} \\ \hline \hline
            \makecell[l]{Social-Trans~\cite{socialtransmotion} \\ ($\alpha=0$)} & \checkmark & & 1.11 & 2.26 \\ \hline
            Ours ($\alpha=0.1$)  & \checkmark & \checkmark & 0.98 & 1.98 \\
            Ours ($\alpha=1.0$)  & \checkmark & \checkmark & 0.96 & 1.95 \\
            Ours ($\alpha=10.0$) & \checkmark & \checkmark & 0.99 & 1.97 \\
            Ours ($\alpha=25.0$) & \checkmark & \checkmark & 0.96 & 1.96 \\
            Ours ($\alpha=100.0$) & \checkmark & \checkmark & 0.97 & 1.91 \\
            Ours ($\alpha=250.0$) & \checkmark & \checkmark & 1.03 & 2.07 \\
            Ours ($\alpha=1,000.0$) & \checkmark & \checkmark & 1.06 & 2.21 \\
            Ours ($\alpha=10,000.0$) & \checkmark & \checkmark & 1.49 & 2.77 \\ \hline
            \makecell[l]{w/o MSE Loss \\ (same as $\alpha\approx\infty$)} &  & \checkmark & 3.52 & 8.20 \\ 
        \end{tabular}
    \end{center}
\end{table}

\section{Additional Results on HTP with EmLoco Loss}
\label{appx:results_JTA}

\subsection{Evaluation on the AMASS Dataset}
\label{sec:amass}
In our framework, as a real-world motion capture dataset with accurate human poses, AMASS~\cite{AMASS} is used as an additional training resource only in stage 1, not in the HTP network training. This raises concerns about unbalanced training resources compared to a baseline that does not use AMASS. From this perspective, to ensure fairness and provide performance on another real-world dataset, we evaluate HTP networks trained only with AMASS.

\paragraph{Experimental setup.}
Following PACER~\cite{TraceandPace}, we split $\sim$200 locomotion sequences in the AMASS dataset into training, validation, and test sets. This split is shared for training the locomotion generator, LocoVal function, and HTP network, \ie, no additional data is introduced for training in the physics simulator. We use global translation and joint positions calculated via forward kinematics from the SMPL parameters in AMASS as trajectory and pose data.

Most training conditions are the same as in the experiments on JTA and JRDB described in the main manuscript. However, since many of the sequences in AMASS are short, we adjust the trajectory length accordingly. Specifically, while JTA and JRDB use $9$ past frames and $12$ future frames at $2.5$ fps, AMASS operates at $30$ fps with $12$ past frames and $30$ future frames.
In addition, the HTP network is trained for 150 epochs, and $\alpha$ for the EmLoco loss weight relative to the GT loss is set to $20$.

\paragraph{Result.}
Table~\ref{tab:result_amass} compares HTP networks~\cite{socialtransmotion} trained only with AMASS. These results show that also in AMASS, which captures real-world human locomotion, the proposed EmLoco loss effectively reduces prediction errors (`Ours w/o filter'). The predictions made by our method improve both the mean ADE / FDE ($4.8\% / 2.0\%$) among $20$ heads and minADE / minFDE ($4.8\% / 3.3\%$), indicating an overall enhancement across multiple predicted trajectories.

Furthermore, applying the LocoVal filter to these predictions (`Ours w/ filter') further improves ADE / FDE while maintaining minADE / minFDE. Given that the ADE / FDE of the rejected trajectories is significantly large, this confirms that the filter successfully eliminates implausible predictions.

Additional results regarding the effect of the LocoVal filter on other datasets are discussed in Sec~\ref{sec:locoval_add}.

\begin{table}[tb]
    \caption{Stochastic HTP results with 20 heads on AMASS~\cite{AMASS}.}
    \label{tab:result_amass}
    \begin{center}
        \scalebox{0.9}{\begin{tabular}{l|cc|cc}
        \textbf{Method} & \textbf{ADE} & \textbf{FDE} & \textbf{minADE} & \textbf{minFDE} \\ \hline \hline
        Social-Trans~\cite{socialtransmotion} & 0.187 & 0.457 & 0.168 & 0.419 \\ \hline 
        \textbf{Ours w/o filter} & 0.178 & 0.448 & \red{0.160} & \red{0.405} \\
        \textbf{Ours w/ filter} & \red{0.175} & \red{0.438} & \red{0.160} & \red{0.405} \\
        Rejected by filter & 0.514 & 1.480 & 0.458 & 1.359 \\
        \end{tabular}}
    \end{center}
\end{table}

\subsection{Detailed Analysis on EmLoco Loss Weight}
\label{sec:emloco_add}

We investigated the significance of the EmLoco loss by varying the hyper-parameter $\alpha$, as defined in Eq.(\ref{eq:mse_and_poseloss}), which controls the balance between the MSE loss and the EmLoco loss. Note that setting $\alpha=0$ is equivalent to excluding the EmLoco loss.  Moreover, we also evaluated the performance when only the EmLoco loss was used as in Table.~\ref{tab:ablation} in the main manuscript. The experimental results are summarized in Table~\ref{tab:hyperparameter}.

As shown in Table~\ref{tab:hyperparameter}, the ADE and FDE remain relatively stable for $\alpha$ values ranging from $0.1$ to $100$. However, when $\alpha$ exceeds $100$, performance begins to degrade. To better understand the balance between the loss components, we examined the scale of each loss function when the training losses converged. On the JTA dataset, the scale of the MSE loss becomes approximately equal to that of the EmLoco loss when $\alpha=1,000$.

These findings indicate that $\alpha$ should be adjusted such that the EmLoco loss remains smaller than the MSE loss.
%
This is because while the EmLoco loss is low around the ground truth future trajectory, no clear peak is observed in the EmLoco loss, unlike the MSE loss.
The widely distributed low values of the EmLoco loss make it difficult to train the HTP network to improve ADE and FDE.
Thus, the MSE loss primarily contributes to reducing ADE and FDE, while the EmLoco loss should be utilized as a support term to enhance plausibility. 

Furthermore, the robustness of the EmLoco loss to changes in the hyperparameter $\alpha$ is evident from the small variations in ADE and FDE observed for $\alpha$ values between $0.1$ and $100$. This robustness allows us to easily integrate the EmLoco loss into the overall loss function.

\subsection{Evaluation of Displacement Errors at Each Timestep}
\label{appx:future}

\begin{figure}[tb]
    \centering
    \includegraphics[width=\linewidth]{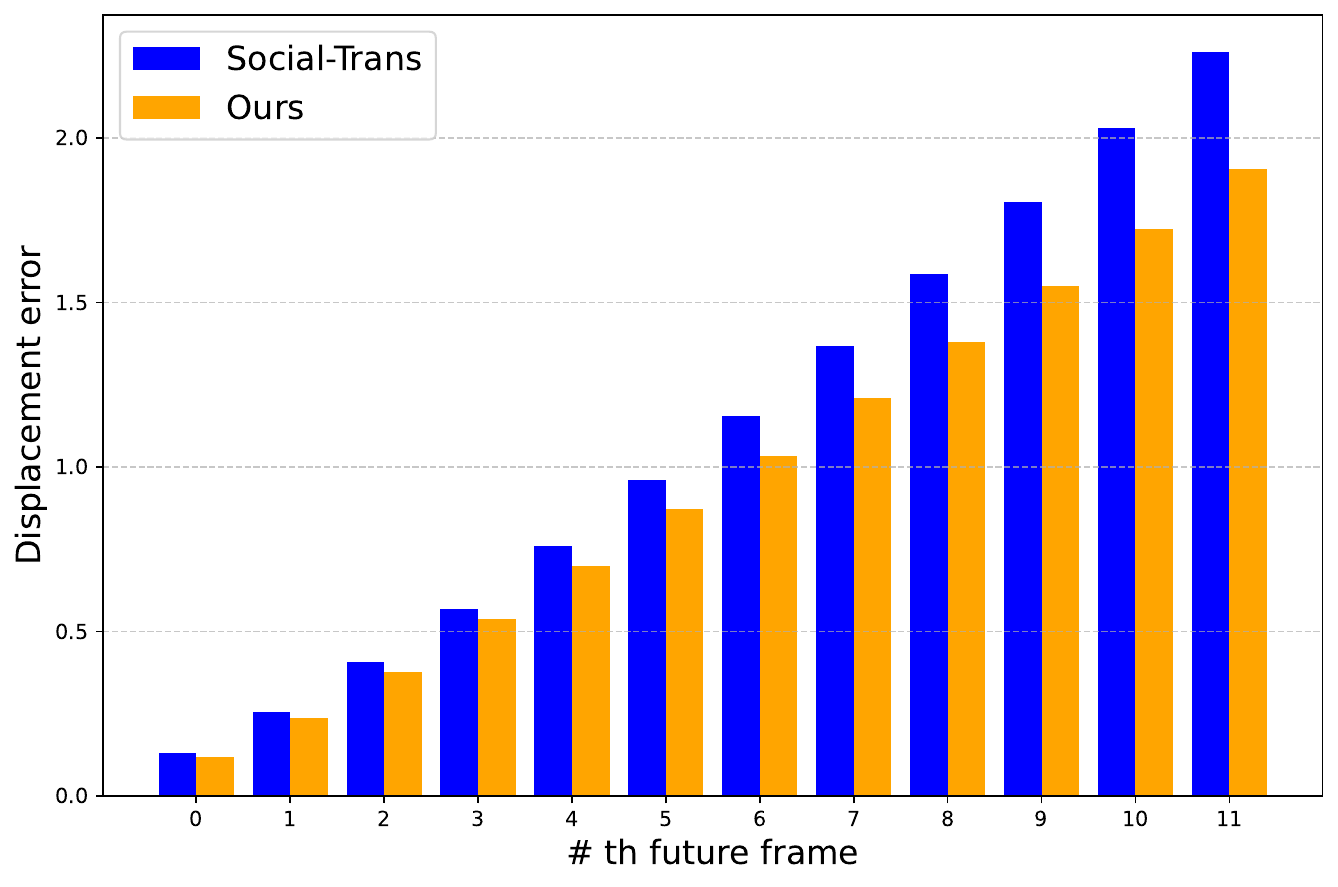}
    \caption{Evaluation of displacement errors for each future timestep.}
    \label{fig:displacement}
\end{figure}

To further demonstrate the effectiveness of the EmLoco loss, Fig.~\ref{fig:displacement} compares the displacement error at each future time step with the baseline. These results correspond to the result on the JTA dataset in Table~\ref{tab:result_tp} of the main paper.

This bar chart highlights two key observations: First, the proposed HTP network trained with the EmLoco loss consistently outperforms the baseline across all time steps. Second, the performance gain relative to the baseline increases toward future frames, with an improvement of $8.8\%$ for the initial frame, growing to $15.7\%$ for the final frame. As also shown in the $\chi^{2}$ distance evaluation in Table~\ref{tab:result_multi} of the main paper, our HTP network acquires physics-based prior knowledge through the EmLoco loss, resulting in features such as velocity that are closer to real-world data. This enhanced ability to capture such features can be the reason for the discrepancy between the baseline and our HTP network accumulating over time.

\subsection{Comparison with MSE Loss Averaging in Stochastic HTP Network Training}
In Sec.~\ref{sec:intro}, we pointed out as follows: `training with the MSE loss alone forces all predicted trajectories to align with a single ground truth. That is, minimizing the MSE essentially reduces the diversity of the predicted trajectories.' Here, we compare the performance of the baseline~\cite{socialtransmotion} trained by this MSE loss averaging with our proposed method.

As shown in Table~\ref{tab:result_meanmse}, while averaging MSE among heads (`meanMSE') improves ADE / FDE, it leads to poorer minADE / minFDE.
This result demonstrates that forcing the outputs of all heads closer to the single ground truth loses diversity.
On the other hand, ours encourages plausibility by EmLoco loss while maintaining diversity by minMSE loss, resulting in better overall performance.

\begin{table}[tb]
    \caption{Stochastic HTP results with 20 heads on JTA. `Deteriminstic' indicates the HTP network with a single head. `minMSE' and `meanMSE' corresponds to the stochastic HTP network trained with the minimum MSE and the mean MSE among heads, respectively.}
    \label{tab:result_meanmse}
    \begin{center}
        \scalebox{0.8}{\begin{tabular}{l|cc|cc}
            \textbf{Method} & \textbf{ADE} & \textbf{FDE} & \textbf{minADE} & \textbf{minFDE} \\ \hline \hline
            Social-Trans (Deterministic) & \red{1.11} & 2.26 & - & - \\ \hline
            Social-Trans (minMSE) & 2.14 & 4.26 & 0.71 & \red{0.54} \\
            Social-Trans (meanMSE) & 1.24 & \red{1.98} & 0.93 & 1.97 \\ \hline
            \textbf{Ours} & 1.80 & 3.56 & \red{0.66} & \red{0.54} \\
        \end{tabular}}
    \end{center}
\end{table}


\subsection{Qualitative Evaluation on the JRDB Dataset}
\label{appx:qualitative}

\begin{figure}[tb]
    \centering
    \includegraphics[width=\linewidth]{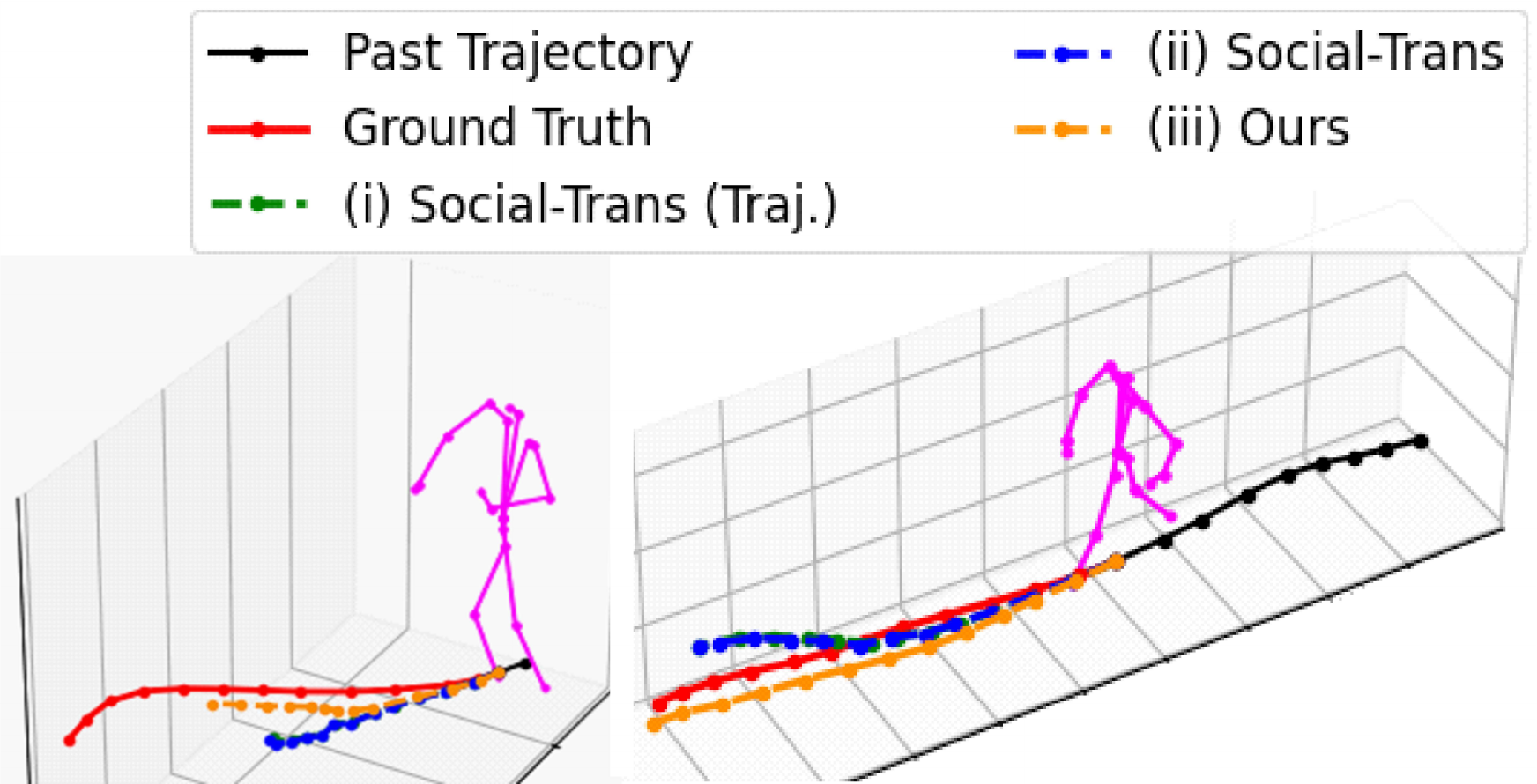}
    \caption{Visualizations of the prediction by the baseline~\cite{socialtransmotion} and our method on JRDB dataset~\cite{JRDB}. Left: results with 2-frame momentary observations. Right: results with 9 frames of observations. `Traj.' indicates that only trajectory is used as input. The scale of \textbf{\textcolor{magenta}{the human pose}} is doubled for a presentation purpose only.}
    \label{fig:jrdb_vis}
\end{figure}

\begin{figure}[tb]
    \centering
    \includegraphics[width=\linewidth]{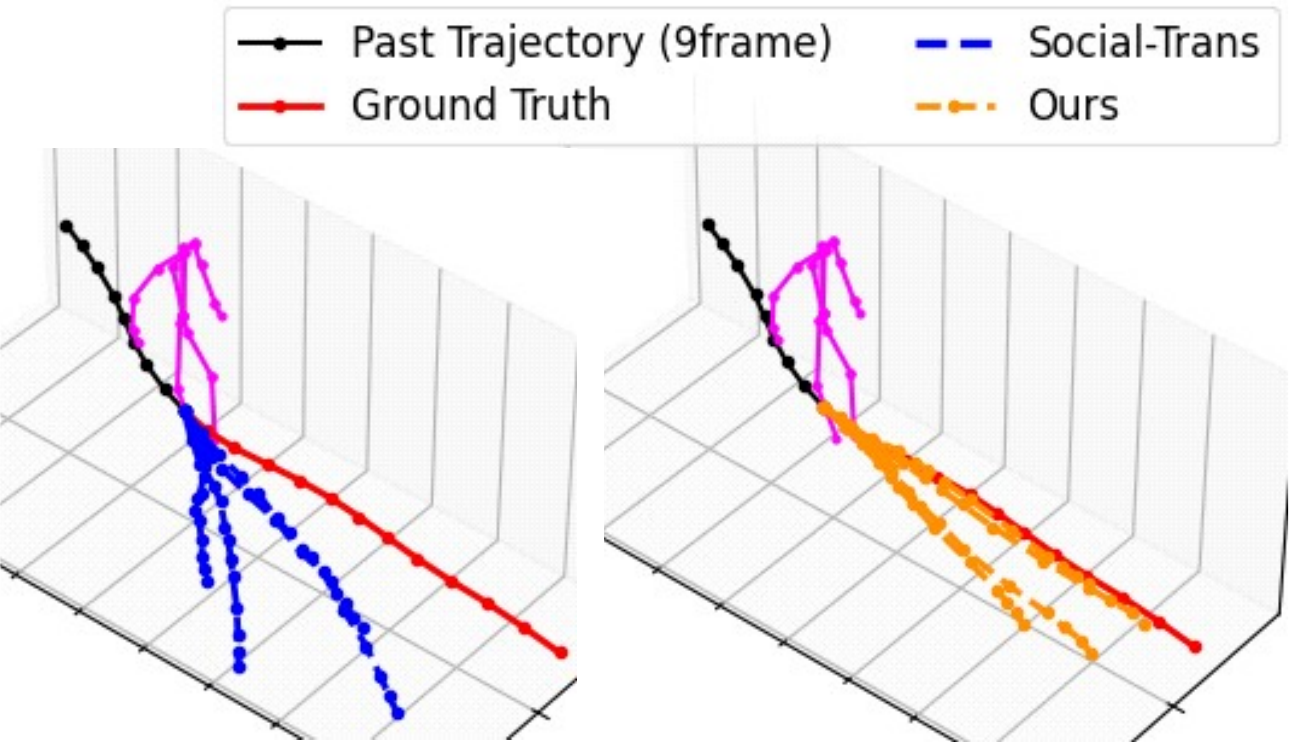}
    \caption{Comparison of stochastic HTP results by the baseline~\cite{socialtransmotion} and ours on the JRDB dataset~\cite{JRDB} with 9 frames of observations. The scale of \textbf{\textcolor{magenta}{the human pose}} is doubled for a presentation purpose only.}
    \label{fig:jrdb_multi}
\end{figure}

Visualizations of the predictions by the baseline~\cite{socialtransmotion} and our method on the JRDB dataset~\cite{JRDB} are shown in Fig.~\ref{fig:jrdb_vis}. Consistent with the results on the JTA dataset~\cite{JTA} (Fig.~\ref{fig:limit_obs} of the main paper), our method predicts plausible trajectories across both observation lengths. Furthermore, the visualization of the stochastic HTP results in Fig.~\ref{fig:jrdb_multi} demonstrates that, while the baseline predictions~\cite{socialtransmotion} deviate significantly from the ground truth for all trajectories, our HTP network successfully predicts plausible trajectories while maintaining reasonable diversity. Even with the imperfect 3D poses in the JRDB dataset, the proposed method makes a prediction close to the ground truth trajectory. These results support the effectiveness of the proposed method in real-world scenarios, not only in simulated environments such as JTA~\cite{JTA}.

\section{Additional Results on LocoVal Filter}
\label{sec:locoval_add}
\subsection{Evaluation with Another HTP Network Incorporating Human Poses}
While we employed the Social-Trans~\cite{socialtransmotion} as the baseline, our method can be applied to other HTP networks. To this end, as an additional baseline incorporating human poses, we evaluate HST. Table~\ref{tab:result_hst} evaluated on the full JRDB dataset (in the main manuscript, a subset provided by the Social-Trans~\cite{socialtransmotion} was used) demonstrates that our LocoVal filter can also improve the ADE / FDE of HST while preserving the minADE / minFDE.

\begin{table}[tb]
    \caption{LocoVal filter on HST~\cite{HST} with $6$ heads on the JRDB dataset~\cite{JRDB}. The filtering threshold $\lambda$ is set to $0.5$.}
    \label{tab:result_hst}
    \begin{center}
        \scalebox{0.8}{\begin{tabular}{l|cc|cc}
            \textbf{Method} & \textbf{ADE} & \textbf{FDE} & \textbf{minADE} & \textbf{minFDE} \\ \hline \hline
            Pretrained HST~\cite{HST} & 0.57 & 0.98 & \red{0.28} & \red{0.45} \\ \hline 
            \textbf{Pretrained HST (w/ filter)} & \red{0.46} & \red{0.80} & \red{0.28} & 0.46 \\ 
            Rejected by filter & 0.95 & 1.64 & 0.60 & 1.01 \\
        \end{tabular}}
    \end{center}
\end{table}

\subsection{Evaluation with HTP Network Based on Neural Social Physics}
There is a type of physics-aware HTP method, called Neural Social Physics (NSP)~\cite{NSP}. While our method evaluates the locomotion of individual persons, NSP~\cite{NSP} models interactions between people through the concept of `social force'. Since these contributions are independent, the proposed method can incorporate NSP and enhance its performance.
Table~\ref{tab:result_nsp} presents the effect of the LocoVal filter with pretrained NSP on the Stanford Drone Dataset (SDD)~\cite{SDD}. Since the trajectory in the SDD dataset is in pixels, we converted the scale to meters for input to the LocoVal function~\cite{opentraj}~\footnote{\url{https://github.com/crowdbotp/OpenTraj/tree/master/datasets/SDD}}. The significantly large ADE / FDE of the rejected samples suggests that even physics-aware existing HTP networks can produce implausible predictions that considerably deviate from the ground truth.

\begin{table}[tb]
    \caption{LocoVal filter on NSP~\cite{NSP} with 20 heads on SDD~\cite{SDD}. The filtering threshold $\lambda$ is set to $0.55$.}
    \label{tab:result_nsp}
    \begin{center}
        \scalebox{0.75}{\begin{tabular}{l|cc|cc}
            \textbf{Method} & \textbf{ADE} & \textbf{FDE} & \textbf{minADE} & \textbf{minFDE}  \\ \hline \hline
            Pretrained NSP~\cite{NSP} & 24.17 & 49.32 & \red{6.52} & \red{10.59} \\ \hline 
            \textbf{Pretrained NSP (w/ filter)} & \red{24.13} & \red{49.24} & \red{6.52} & \red{10.59} \\
            Rejected by filter & 256.24 & 548.09 & 210.39 & 464.17 \\
        \end{tabular}}
    \end{center}
\end{table}

\subsection{Detailed Analysis on Filtering Threshold}
\label{appx:threshold}

The LocoVal filter introduced in Sec.~\ref{method:filtering} of the main paper allows control over the strictness of the trajectory filtering by changing the threshold $\lambda$. While Tables~\ref{tab:result_filter} and~\ref{tab:ethucy} in the main paper present results for a single threshold, this section investigates how the filtering results vary with different threshold settings. In addition to the results on the JTA~\cite{JTA} and ETH / UCY datasets~\cite{ETH,UCY} presented in the main manuscript, this section also provides the filtering results on the JRDB dataset~\cite{JRDB}. 
Since the JRDB dataset lacks 3D poses for some pedestrians, we evaluate the performance both before and after filtering for pedestrians with 3D poses.

Tables~\ref{tab:filter_jta_appx},~\ref{tab:filter_jrdb_appx}, and~\ref{tab:filter_eth_appx} presents the results on the JTA, JRDB, and ETH / UCY datasets.

Consistent with the experimental results in the main text, the LocoVal filter improved the average ADE / FDE across a wide range of settings, including different datasets, varying numbers of prediction heads, different numbers of input frames, and HTP networks trained with and without the EmLoco loss. Interestingly, however, a certain trade-off can be observed in these results.
When the threshold $\lambda$ is relaxed (See $\lambda=0.65, 0.70, 0.75$ in the JTA, JRDB, and ETH / UCY datasets, respectively), it does not significantly impact the filtering performance.
For example, Table~\ref{tab:filter_jta_appx} shows that the ADE with 9 frames of observation degrades from 1.81/1.97 ($\lambda$ = 0.70) to 1.83/2.02 ($\lambda$ = 0.65) with Social-Trans, and from 1.65/1.76 ($\lambda$ = 0.70) to 1.66/1.78 ($\lambda$ = 0.65) with our method. This performance degradation is quite small.
However, it becomes capable of rejecting extremely incorrect results (\eg, ADE and FDE of rejected samples on the ETH / UCY dataset are $14.48$ and $32.87$). Setting the high threshold (See $\lambda=0.70, 0.75, 0.80$ in the JTA, JRDB, and ETH / UCY datasets, respectively) sometimes degrades the performance (\eg, after the filtering with $\lambda=0.80$ on the JRDB dataset with momentary observations, ADE / FDE of our HTP network become worse than those of without filtering). This is because high plausibility is not always equivalent to being close to the ground truth trajectory. While feasible locomotion has a certain level of plausibility, humans may perform implausible locomotion due to interactions with obstacles or others. While the threshold $\lambda$ can be freely controlled, it is necessary to set it appropriately, considering this trade-off. 

\begin{table*}[tb]
\caption{Results of the LocoVal filter with various $\lambda$ for stochastic HTP on the JTA~\cite{JTA} dataset with 9 and 2-frame observations. The results are presented as left/right values, where the left denotes evaluations with $5$ heads and the right denotes $20$ heads. 
}
\label{tab:filter_jta_appx}
\centering
\scalebox{1.1}{
\begin{tabular}{cl|c|cc|cc}
\multicolumn{2}{c|}{\multirow{2}{*}{\textbf{Method}}} & \multirow{2}{*}{$\lambda$} & \multicolumn{2}{c|}{\textbf{9 frames}} & \multicolumn{2}{c}{\textbf{2 frames}} \\ \cline{4-7} 
\multicolumn{2}{c|}{}                                 & & \multicolumn{1}{c}{ADE}     & FDE      & \multicolumn{1}{c}{ADE}       & FDE        \\ \hline
\multicolumn{1}{c}{\multirow[c]{7}{*}{\rotatebox{90}{\makecell{Social-Trans \\ (w/o $\mathcal{L}_\text{E}$)}}}} 
& w/o Filtering & - & 1.86/2.14 & 3.51/4.26 & 2.23/2.46 & 4.55/5.07 \\ \cdashline{2-7}
& w/ Filtering & 0.65 & 1.83/2.02 & 3.48/4.02 & 2.10/2.28 & 4.34/4.70 \\ 
& w/ Filtering & 0.70  & 1.81/1.97 & 3.49/3.93 & 2.06/2.21 & 4.27/4.56 \\
& w/ Filtering & 0.75 & 1.85/1.94 & 3.69/3.97 & 2.11/2.17 & 4.46/4.56 \\ \cdashline{2-7}
& Rejected & 0.65     & 2.71/3.99 & 4.24/7.88 & 4.92/5.09 & 8.91/10.52 \\
& Rejected & 0.70      & 2.37/3.41 & 3.75/6.65 & 3.83/4.32 & 7.16/8.87 \\
& Rejected & 0.75      & 1.92/2.50 & 3.19/4.78 & 2.50/3.03 & 4.76/6.06 \\ \hline
\multicolumn{1}{c}{\multirow{7}{*}{\rotatebox{90}{\makecell[c]{\textbf{Ours} \\ \textbf{(w/ $\mathcal{L}_\text{E}$)}}}}}         
& w/o Filtering & - & 1.68/1.80 & 3.34/3.56 & 1.94/2.12 & 3.96/4.47  \\ \cdashline{2-7}
& w/ Filtering & 0.65 & 1.66/1.78             & 3.32/3.54             & 1.92/2.10             & 3.93/4.44  \\ 
& w/ Filtering & 0.70  & 1.65/1.76 & 3.31/3.52 & 1.90/2.08 & 3.88/4.41  \\ 
& w/ Filtering & 0.75 & 1.64/1.75 & 3.42/3.62 & 1.90/2.08 & 4.00/4.50  \\ \cdashline{2-7}
& Rejected & 0.65     & 2.68/2.81 & 4.41/4.89 & 3.80/4.15 & 7.11/8.02  \\
& Rejected & 0.70      & 2.32/2.40 & 3.86/4.19 & 3.22/3.25 & 6.18/6.41  \\
& Rejected & 0.75     & 1.83/1.93 & 3.11/3.44 & 2.09/2.24 & 3.83/4.36  \\
\end{tabular}
}
\end{table*}

\begin{table*}[tb]
\caption{Results of the LocoVal filter with various $\lambda$ for stochastic HTP on the JRDB~\cite{JRDB} dataset with 9 and 2-frame observations. The results are presented as left/right values, where the left denotes evaluations with $5$ heads and the right denotes $20$ heads. 
}
\label{tab:filter_jrdb_appx}
\centering
\scalebox{1.1}{
\begin{tabular}{cl|c|cc|cc}
\multicolumn{2}{c|}{\multirow{2}{*}{\textbf{Method}}} & \multirow{2}{*}{$\lambda$} & \multicolumn{2}{c|}{\textbf{9 frames}} & \multicolumn{2}{c}{\textbf{2 frames}} \\ \cline{4-7} 
\multicolumn{2}{c|}{}                                 & & \multicolumn{1}{c}{ADE}     & FDE      & \multicolumn{1}{c}{ADE}       & FDE        \\ \hline
\multicolumn{1}{c}{\multirow[c]{7}{*}{\rotatebox{90}{\makecell{Social-Trans \\ (w/o $\mathcal{L}_\text{E}$)}}}} 
& w/o Filtering       & - & 0.71/0.76 & 1.40/1.58 & 0.63/0.68 & 1.28/1.46  \\ \cdashline{2-7}
& w/ Filtering & 0.70  & 0.70/0.73 & 1.39/1.52 & 0.59/0.67 & 1.21/1.43 \\ 
& w/ Filtering & 0.75  & 0.69/0.71 & 1.38/1.48 & 0.58/0.66 & 1.18/1.42 \\ 
& w/ Filtering & 0.80  & 0.68/0.64 & 1.36/1.34 & 0.56/0.60 & 1.13/1.30 \\ \cdashline{2-7}
& Rejected & 0.70     & 1.60/2.31 & 3.03/4.73 & 1.86/1.60 & 4.10/3.41  \\
& Rejected & 0.75     & 1.25/1.62 & 2.37/3.30 & 1.44/1.13 & 3.12/2.41  \\
& Rejected & 0.80     & 0.75/0.89 & 1.48/1.83 & 0.76/0.77 & 1.55/1.64  \\ \hline
\multicolumn{1}{c}{\multirow{7}{*}{\rotatebox{90}{\makecell[c]{\textbf{Ours} \\ \textbf{(w/ $\mathcal{L}_\text{E}$)}}}}}         
& w/o Filtering       & - & 0.61/0.71 & 1.26/1.50 & 0.57/0.67 & 1.17/1.46  \\ \cdashline{2-7}
& w/ Filtering & 0.70  & 0.60/0.70 & 1.23/1.49 & 0.56/0.65 & 1.15/1.43  \\
& w/ Filtering & 0.75 & 0.59/0.69 & 1.21/1.46 & 0.57/0.64 & 1.16/1.40  \\ 
& w/ Filtering & 0.80  & 0.59/0.64 & 1.22/1.38 & 0.59/0.61 & 1.20/1.33  \\ \cdashline{2-7}
& Rejected & 0.70      & 1.76/1.60 & 3.42/3.20 & 0.97/1.51 & 1.97/3.25  \\
& Rejected & 0.75     & 1.15/1.19 & 2.32/2.45 & 0.74/1.14 & 1.52/2.47  \\
& Rejected & 0.80      & 0.66/0.79 & 1.36/1.64 & 0.57/0.73 & 1.16/1.59  \\
\end{tabular}
}
\end{table*}

\begin{table*}[tb]
\centering
\caption{Results of zero-shot filtering with various $\lambda$ by the LocoVal filter on the predictions of a pre-trained 20 heads trajectory predictor~\cite{Eqmotion} on the ETH / UCY dataset~\cite{ETH,UCY}. `Mean' represents the average performance across the $5$ subsets.}
\scalebox{0.93}{
\begin{tabular}{cl|c|cc:cc:cc:cc:cc|cc}
\multicolumn{2}{c|}{\multirow{2}{*}{\textbf{Method}}} & \multirow{2}{*}{$\lambda$} & \multicolumn{2}{c:}{\textbf{ETH}} & \multicolumn{2}{c:}{\textbf{HOTEL}} & \multicolumn{2}{c:}{\textbf{UNIV}} & \multicolumn{2}{c:}{\textbf{ZARA1}} & \multicolumn{2}{c|}{\textbf{ZARA2}} & \multicolumn{2}{c}{\textbf{Mean}} \\ \cline{4-15} 

\multicolumn{2}{c|}{} & & \multicolumn{1}{c}{ADE} & \multicolumn{1}{c:}{FDE} & \multicolumn{1}{c}{ADE} & \multicolumn{1}{c:}{FDE} & \multicolumn{1}{c}{ADE} & \multicolumn{1}{c:}{FDE} & \multicolumn{1}{c}{ADE} & \multicolumn{1}{c:}{FDE} & \multicolumn{1}{c}{ADE} & \multicolumn{1}{c|}{FDE} & \multicolumn{1}{c}{ADE} & \multicolumn{1}{c}{FDE} \\ \hline\hline

\multicolumn{1}{c}{\multirow[c]{7}{*}{\rotatebox{90}{\makecell{Pretrained \\ EqMotion~\cite{Eqmotion}}}}} 
& w/o Filtering  & - & 2.18 & 4.63 & 0.64 & 1.31 & 1.30 & 2.81 & 0.82 & 1.84 & 0.65 & 1.47 & 1.12 & 2.41 \\ \cdashline{2-15}
& w/ Filtering & 0.75 & 1.58 & 3.27 & 0.63 & 1.30 & 1.05 & 2.26 & 0.82 & 1.84 & 0.65 & 1.47 & 0.95 & 2.03 \\
& w/ Filtering & 0.80 & 1.41 & 2.88 & 0.61 & 1.26 & 0.93 & 2.04 & 0.80 & 1.80 & 0.64 & 1.45 & 0.88 & 1.89 \\
& w/ Filtering & 0.85 & 2.11 & 4.62 & 0.81 & 1.76 & 1.30 & 2.84 & 0.77 & 1.70 & 0.86 & 1.95 & 1.17 & 2.57 \\ \cdashline{2-15}
& Rejected & 0.75 & 14.48 & 32.87 & 5.46 & 11.12 & 7.68 & 16.57 & 4.50 & 8.81 & 4.28 & 8.77 & 7.28 & 15.63 \\
& Rejected & 0.80 & 8.89 & 19.72 & 2.69 & 5.53 & 4.33 & 9.18 & 1.70 & 3.67 & 2.21 & 4.72 & 3.96 & 8.56 \\
& Rejected & 0.85 & 2.18 & 4.63 & 0.64 & 1.31 & 1.30 & 2.81 & 0.82 & 1.84 & 0.65 & 1.47 & 1.12 & 2.41 \\

\end{tabular}
}
\label{tab:filter_eth_appx}
\end{table*}

\subsection{Trade-off between Retaining Plausible Samples and Suppressing Others}
\begin{figure}[tb]
  \centering
  \includegraphics[width=\linewidth]{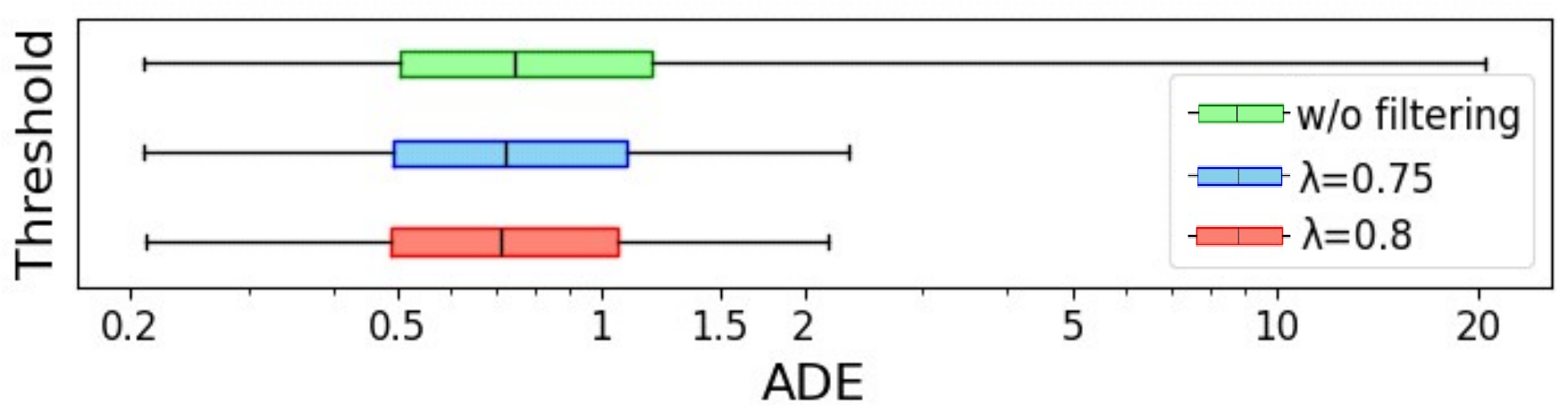}
  \caption{Box plot of ADE before/after LocoVal filtering on EqMotion~\cite{Eqmotion} with 20 heads on ETH / UCY~\cite{ETH, UCY}. $\lambda$ is the threshold.}
  \label{fig:box-plot}
\end{figure}

While our LocoVal filter can suppress implausible trajectories, trajectories close to the ground truth still can be judged as implausible. This is the potential trade-off of the LocoVal filter between retaining plausible samples and suppressing others. However, according to the results shown in the box plot in Fig.~\ref{fig:box-plot}, our LocoVal filter preserves the minimum and median well while effectively rejecting trajectories with high ADE, achieving its purpose.

\subsection{Detailed Analysis on Evaluated Plausibility Scores}
\label{appx:locoval}
\begin{figure}[tb]
  \begin{minipage}[tb]{\columnwidth}
    \centering
    \includegraphics[width=\columnwidth]{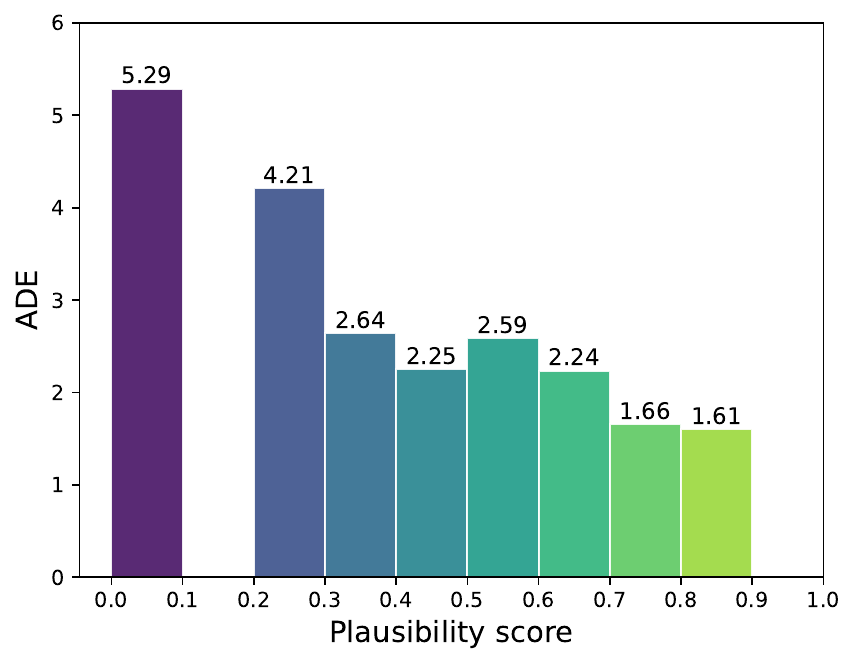}
    \subcaption{ADE of our HTP network vs. plausibility scores.}
  \end{minipage}
  \begin{minipage}[tb]{\columnwidth}
    \centering
    \includegraphics[width=\columnwidth]{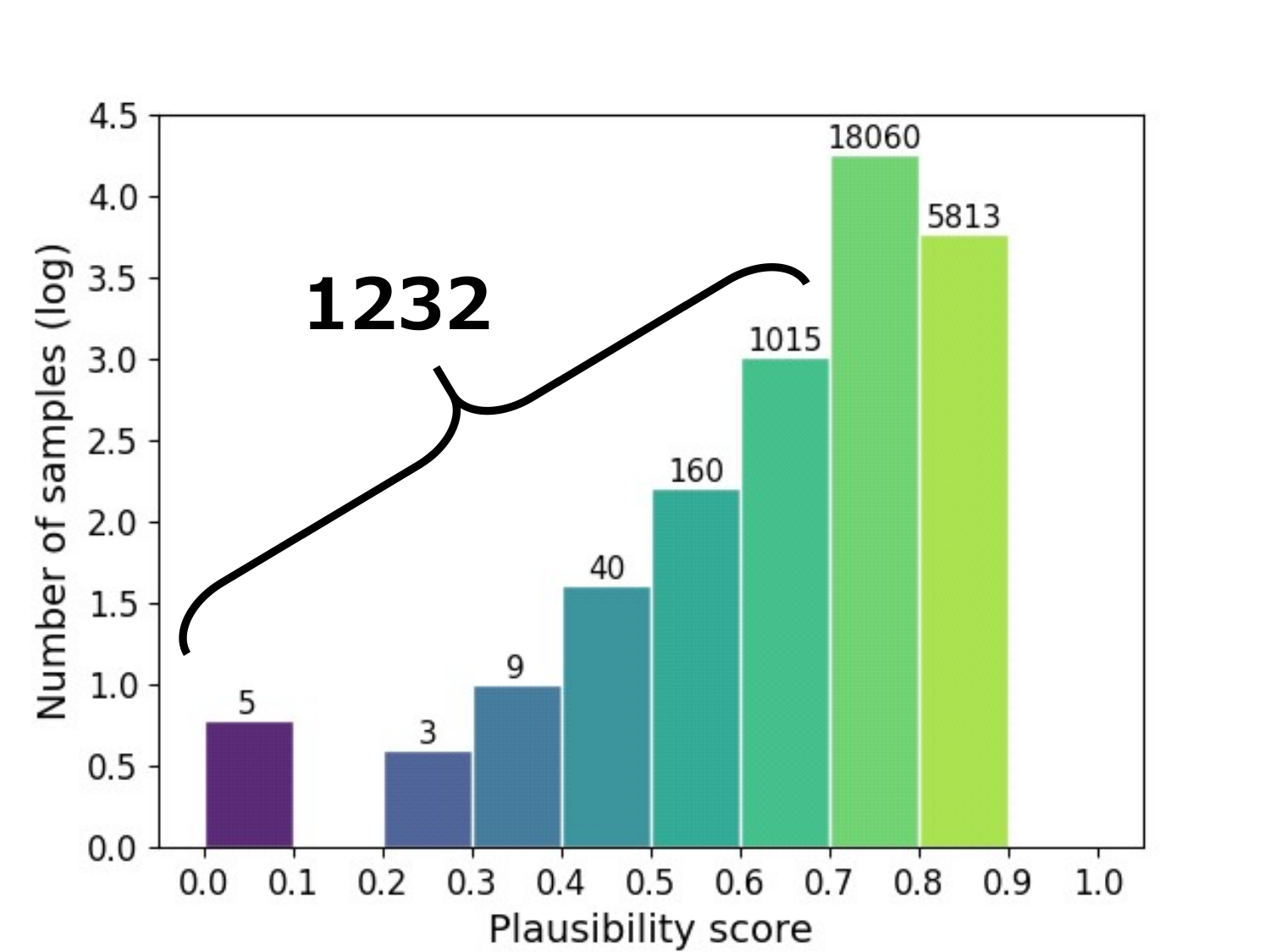}
    \subcaption{Number of samples vs. plausibility scores.}
  \end{minipage}
  \caption{Bar charts illustrating the plausibility scores evaluated by the LocoVal function and the corresponding bin-wise ADE and the number of samples. The numbers displayed above each bin indicate the ADE and the count of predicted trajectories with the corresponding plausibility score. The results show the predictions made by $5$-head our HTP network with $9$ frames observations on the JTA dataset.}
  \label{fig:value_ours}
\end{figure}

\begin{figure}[tb]
  \begin{minipage}[tb]{\columnwidth}
    \centering
    \includegraphics[width=\columnwidth]{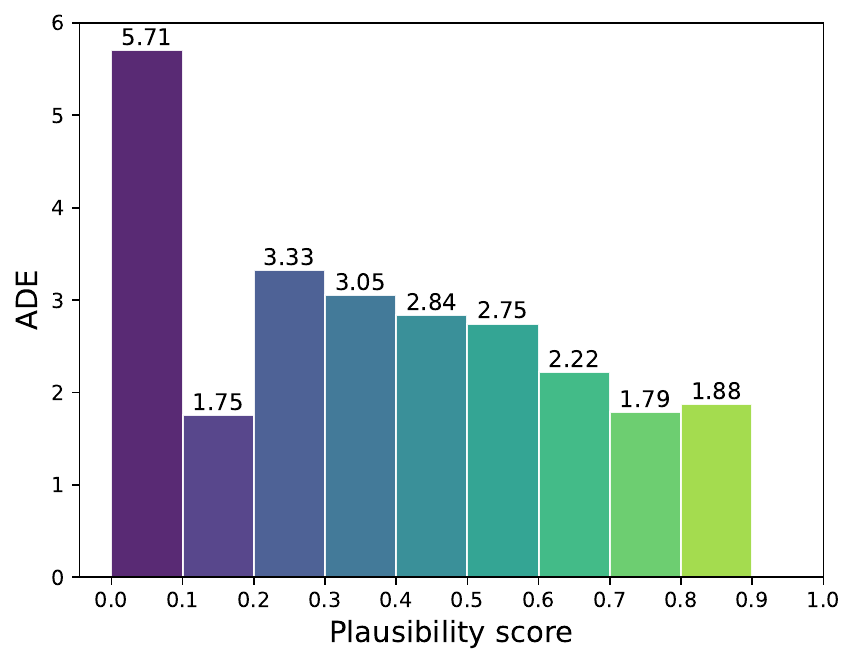}
    \subcaption{ADE of the basline~\cite{socialtransmotion} vs. plausibility scores.}
  \end{minipage}
  \begin{minipage}[tb]{\columnwidth}
    \centering
    \includegraphics[width=\columnwidth]{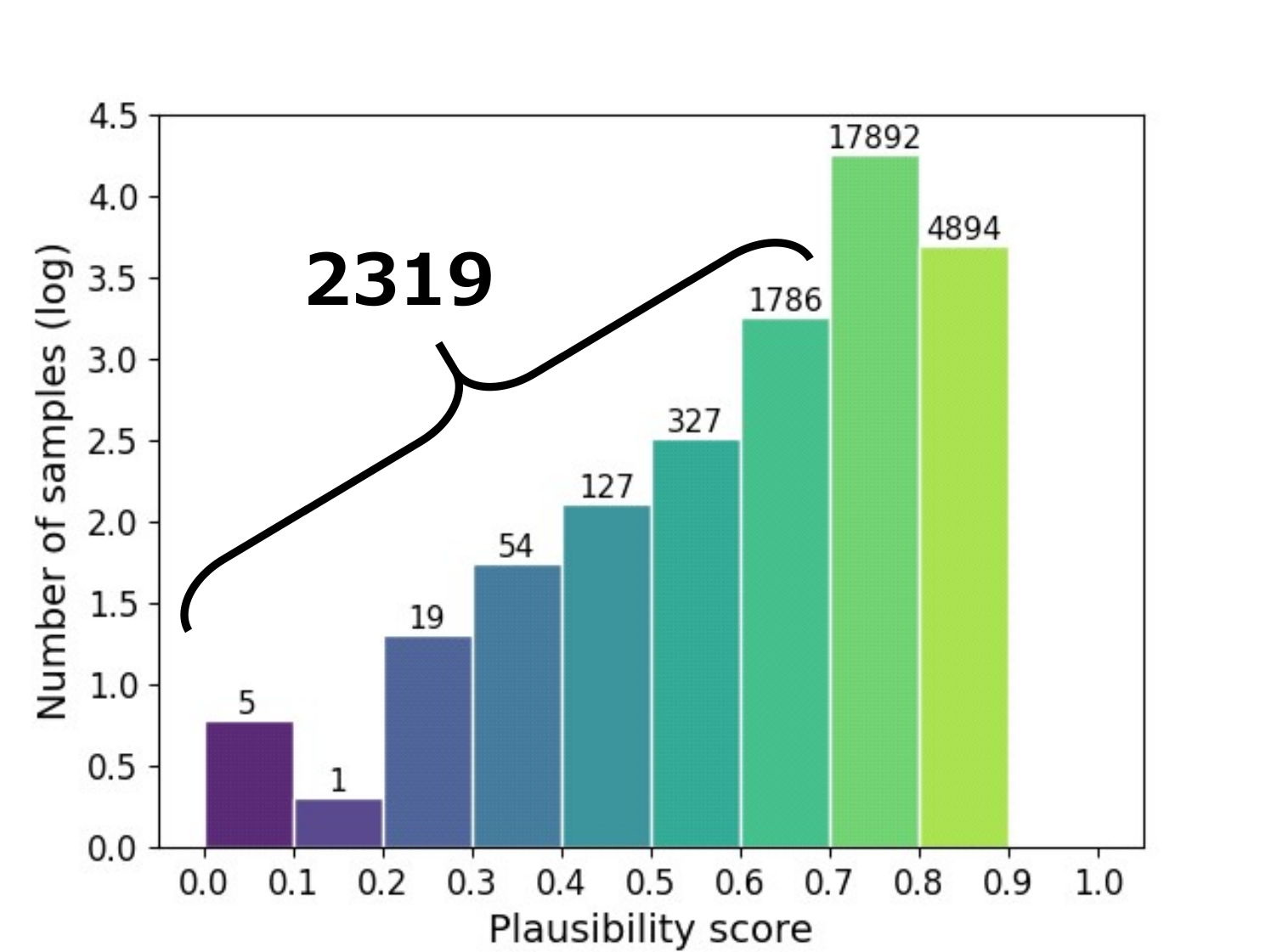}
    \subcaption{Number of samples vs. plausibility scores.}
  \end{minipage}
  \caption{Bar charts illustrating the plausibility scores evaluated by the LocoVal function and the corresponding bin-wise ADE and the number of samples. The numbers displayed above each bin indicate the ADE and the count of predicted trajectories with the corresponding plausibility score. The results show the predictions made by $5$-head Social-Trans~\cite{socialtransmotion} with $9$ frames observations on the JTA dataset.}
  \label{fig:value_baseline}
\end{figure}

The performance improvement by the LocoVal filter depends on the quality of the plausibility score evaluation performed by the LocoVal function, which is trained by our proposed method. To investigate this, we provide bar charts regarding the plausibility scores evaluated by our LocoVal function and the corresponding HTP performance in Figs.~\ref{fig:value_ours} and~\ref{fig:value_baseline}. Figure~\ref{fig:value_ours} presents the results with our HTP network and Fig.~\ref{fig:value_baseline} represents the results of the baseline~\cite{socialtransmotion}. While high plausibility is not always equivalent to closeness to the ground truth trajectory as mentioned above, both figures demonstrate a consistent trend: trajectories with lower plausibility scores exhibit higher ADE values, while those with higher scores tend to have lower ADE values. This observation indicates that trajectories with higher plausibility scores are more likely to be closer to the ground truth trajectories. These results support the validity of our proposed LocoVal function, which has been learned to effectively evaluate plausibility as embodied locomotion.

Furthermore, comparisons of the plausibility scores between the trajectories predicted by our HTP network and those predicted by the baseline are shown in Fig.~\ref{fig:value_ours} and Fig.~\ref{fig:value_baseline}, respectively. These results reveal that the baseline, Social-Trans~\cite{socialtransmotion}, generates a higher proportion of trajectories with lower plausibility scores. Comparing the number of trajectories with a plausibility score of $0.7$ or lower, the baseline has $2,319$ trajectories, whereas the proposed method reduces this to $1,232$, achieving a $46.9\%$ decrease. This indicates that the Social-Trans, which is trained solely on the MSE with respect to the ground truth, is unable to generate plausible trajectories. In contrast, by incorporating the EmLoco loss into the training objective, our model is capable of predicting more plausible trajectories.

\subsection{Visualization of Plausibility Scores}
\label{appx:score_vis}
\begin{figure}[tb]
    \centering
    \includegraphics[width=\linewidth]{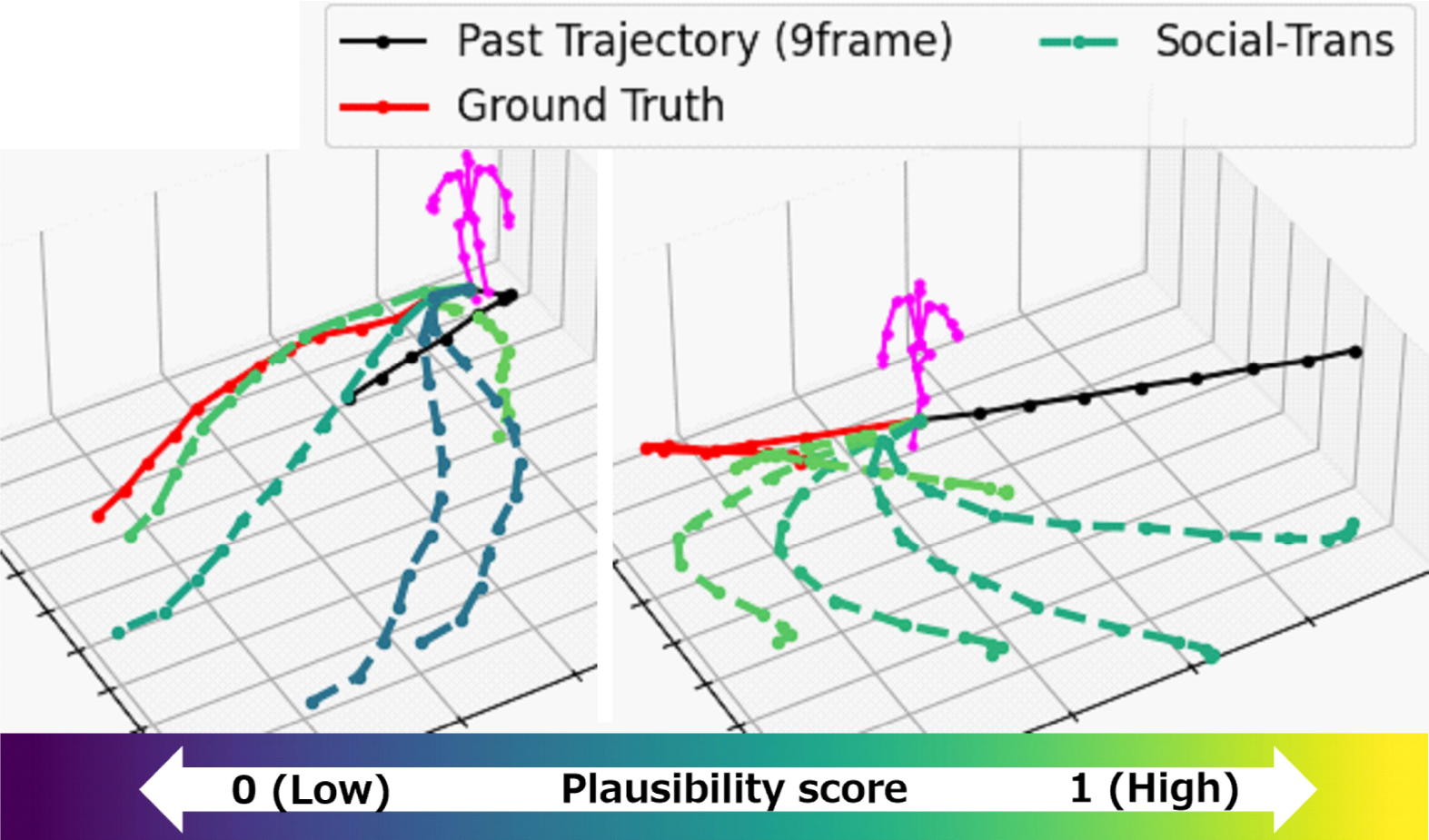}
    \caption{Plausibility score evaluation by our LocoVal function on the JTA dataset with 9 frames of observations. The scale of \textbf{\textcolor{magenta}{the human pose}} is doubled for a presentation purpose only.}
    \label{fig:score_good}
\end{figure}

We provide qualitative examples of plausibility score evaluation by our LocoVal function in Fig.~\ref{fig:score_good}. Here, to evaluate the validity of the plausibility score evaluation by the LocoVal function, we utilize predictions from Social-Trans~\cite{socialtransmotion}, which predicts diverse trajectories ranging from plausible to implausible. In the left case, the two trajectories indicated in cyan are implausible due to their excessive movement from the current pose. As expected, these trajectories have lower plausibility scores and are farther from the ground truth trajectory. In the right case, predicted trajectories that deviate from the ground truth and involve sharper turns tend to have lower plausibility scores. Our LocoVal filter enables more plausible and accurate HTP by excluding such implausible trajectories at inference.

\subsection{Visualization of LocoVal Filtering on ETH / UCY Datasets}
\label{appx:eth_vis}
\begin{figure}[tb]
    \centering
    \includegraphics[width=\linewidth]{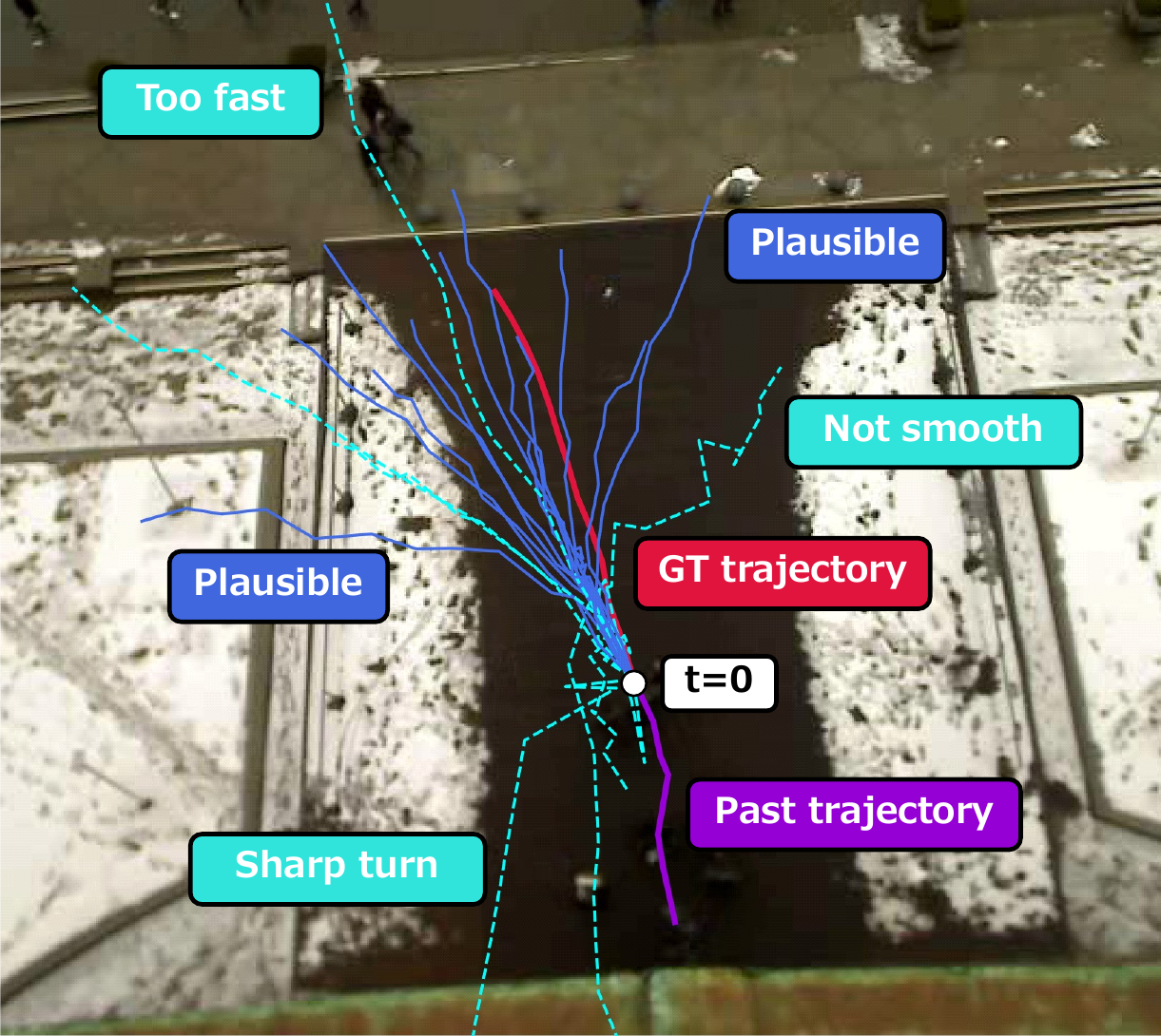}
    \caption{HTP result of pre-trained EqMotion~\cite{Eqmotion} with our LocoVal filter on ETH / UCY dataset~\cite{ETH,UCY}. \blue{Blue lines} represent filtered trajectories and \textcolor{cyan}{\textbf{light blue dashed line}} represent rejected trajectories.}
    \label{fig:eth_filtering}
\end{figure}

Furthermore, we visualized the effect of the LocoVal filter with $\lambda=0.8$ on the predictions of the pre-trained EqMotion~\cite{Eqmotion} on the ETH / UCY dataset~\cite{ETH,UCY}, as shown in Fig.~\ref{fig:eth_filtering}. This demonstrates that, despite performing zero-shot filtering using the LocoVal function that does not rely on pose information, the LocoVal filter effectively identifies and eliminates implausible trajectories (\eg, too fast, involving sharp turns, or lack of smoothness). Again, the filtered results maintain trajectory diversity while constraining predictions to those plausible and closer to the ground truth, confirming the effectiveness of the trained LocoVal function at inference.

\section{Limitations and Future Work}
\label{appx:failure}

\begin{figure}[tb]
    \centering
    \includegraphics[width=\linewidth]{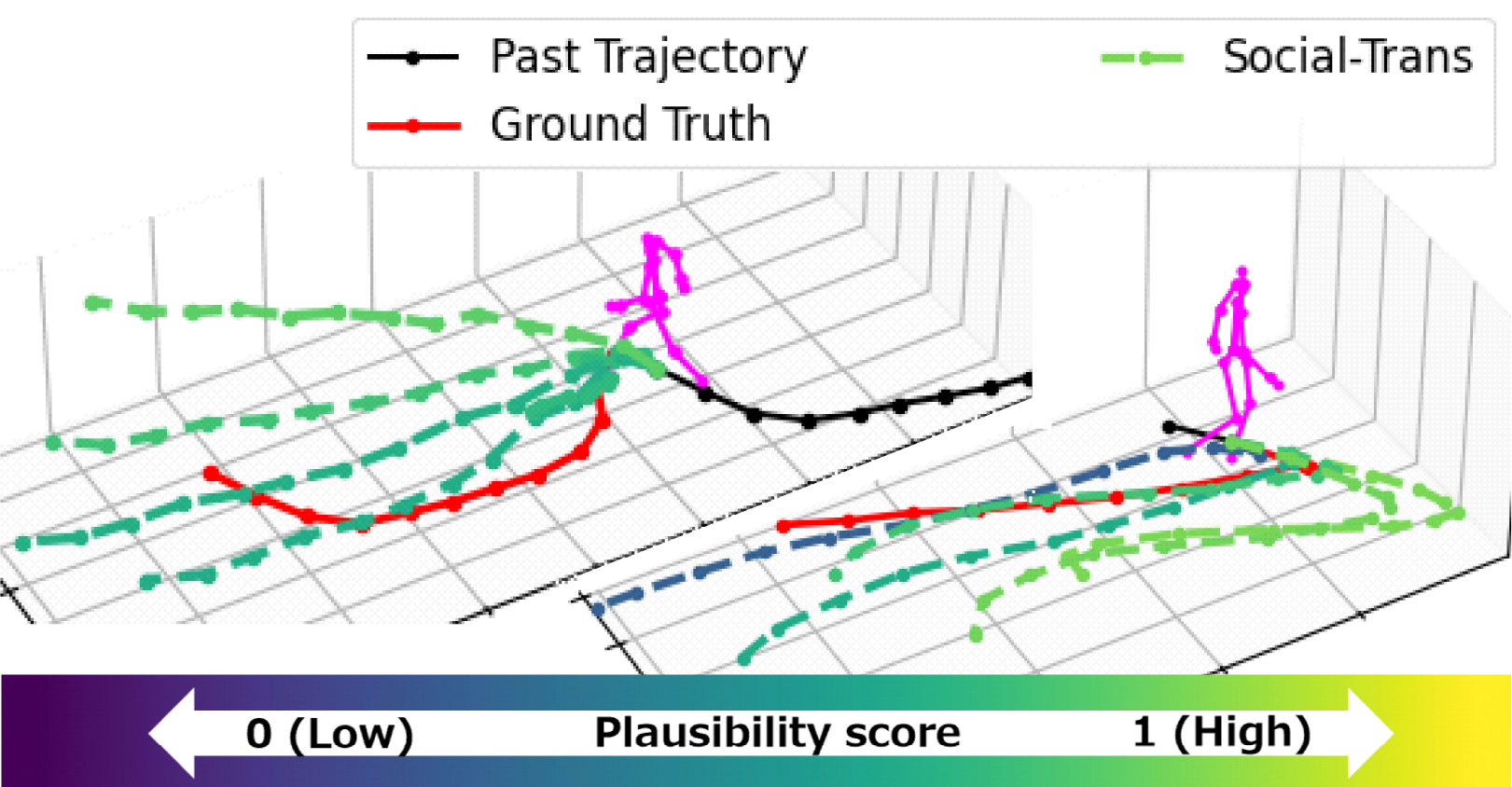}
    \caption{Failure cases and estimated plausibility scores on the JTA dataset. The scale of \textbf{\textcolor{magenta}{the human pose}} is doubled for a presentation purpose only.}
    \label{fig:score_failure}
\end{figure}

While the proposed method improves ADE and FDE for many samples, there are still some failure cases. One mode of such a case is shown in Fig.~\ref{fig:score_failure}. Since the JTA dataset~\cite{JTA} is synthetic, some of the ground truth trajectories are implausible. Consequently, as illustrated in Fig.~\ref{fig:score_failure}, there are cases where trajectories with lower plausibility scores are closer to the ground truth among multiple predicted trajectories while our LocoVal function reasonably scores each trajectory. As mentioned earlier, people would move along such implausible trajectories in real-world scenarios due to interactions with others. Therefore, future research may integrate techniques to score trajectories considering surrounding contexts~\cite{stimulus}.

\begin{figure}[tb]
    \centering
    \includegraphics[width=\linewidth]{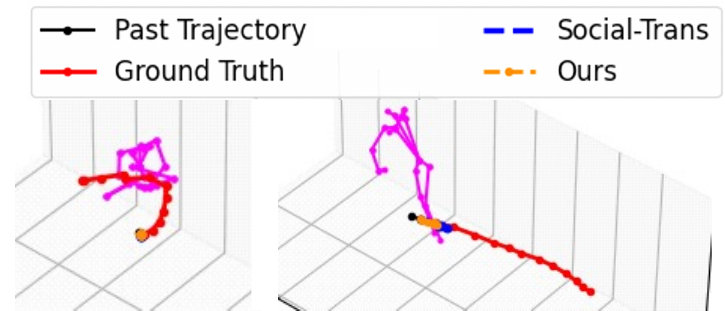}
    \caption{Failure cases in the JRDB dataset. The scale of \textbf{\textcolor{magenta}{the human pose}} is doubled for a presentation purpose only.}
    \label{fig:jrdb_failure}
\end{figure}

Furthermore, as mentioned in Sec.~\ref{sec:discussion} of the main manuscript, the JRDB dataset~\cite{JRDB} contains incorrect 3D poses. Although our experiments demonstrate that our proposed method can improve HTP performance even with such imperfect poses, as shown in Fig.~\ref{fig:jrdb_failure}, both the Social-Trans~\cite{socialtransmotion} and our proposed method make significant errors when faced with incorrect poses. As discussed in the main paper, it is expected that more accurate 3D pose estimation or methods that can account for pose uncertainty~\cite{VL4Pose,VATL4Pose} can further enhance the effectiveness of our proposed method in real-world scenarios.

Lastly, while training with SMPL humanoid~\cite{SMPL} and Isaac Gym~\cite{isaac} enables realistic human motion generation~\cite{Physdiff} and accurate reconstruction of real-world human locomotion~\cite{HumanDynamicsAD}, the more accurate physics simulators are anticipated to more realistically simulate the human locomotion in the real world. 

\end{document}